\documentclass[10pt,journal,compsoc]{IEEEtran}
\usepackage{algorithmic}
\usepackage{algorithm}
\usepackage{array}
\usepackage[caption=false,font=normalsize,labelfont=sf,textfont=sf]{subfig}
\usepackage{textcomp}
\usepackage{stfloats}
\usepackage{url}
\usepackage{verbatim}
\usepackage{graphicx}
\usepackage{amsmath,amssymb,amsfonts}
\usepackage{epsfig}
\usepackage{dsfont}
\usepackage{color}
\usepackage{boldline,multirow }
\usepackage{siunitx}
\usepackage{booktabs}
\usepackage{soul}
\usepackage{cite}
\usepackage{cases}
\usepackage{hyperref}
\usepackage{bbding}
\usepackage{makecell}
\begin{document}

\title{MoMBS: Mixed-order minibatch sampling enhances model training from diverse-quality images}

\author{Han~Li,~Hu~Han,~\IEEEmembership{Member, IEEE}, and~S.~Kevin~Zhou, \IEEEmembership{Fellow, IEEE}}

\markboth{Journal of \LaTeX\ Class Files,~Vol.~14, No.~8, 2023}%
{Shell \MakeLowercase{\textit{et al.}}: Bare Advanced Demo of IEEEtran.cls for IEEE Computer Society Journals}

\IEEEtitleabstractindextext{%
\begin{abstract}
  Natural images exhibit label diversity (clean vs. noisy) in noisy-labeled image classification and prevalence diversity (abundant vs. sparse) in long-tailed image classification. Similarly, medical images in universal lesion detection (ULD) exhibit substantial variations in image quality, encompassing attributes such as clarity and label correctness.
  How to effectively leverage training images with diverse qualities becomes a problem in learning deep models. Conventional training mechanisms, such as self-paced curriculum learning (SCL) and online hard example mining (OHEM), relieve this problem by reweighting images with high loss values. Despite their success, these methods still confront two challenges: (i) the loss-based measure of sample hardness is imprecise, preventing optimum handling of different cases, and (ii) there exists under-utilization in SCL or over-utilization OHEM with the identified hard samples. To address these issues, this paper revisits the minibatch sampling (MBS), a technique widely used in deep network training but largely unexplored concerning the handling of diverse-quality training samples. We discover that the samples within a minibatch influence each other during training; thus, we propose a novel Mixed-order Minibatch Sampling (MoMBS) method to optimize the use of training samples with diverse qualities. MoMBS introduces a measure that takes both loss and uncertainty into account to surpass a sole reliance on loss and allows for a more refined categorization of high-loss samples by distinguishing them as either poorly labeled and under represented or well represented and overfitted. We prioritize under represented samples as the main gradient contributors in a minibatch and keep them from the negative influences of poorly labeled or overfitted samples with a mixed-order minibatch sampling design. Our approach leads to a more precise measurement of sample difficulty, preventing an indiscriminative treatment for under- or over-utilization of hard samples. We conduct extensive experimental evaluations to validate the performance and generalization ability of our method with four tasks including ULD on DeepLesion dataset, COVID segmentation on Seg-19 dataset, long-tailed image classification on CIFAR100-LT, and noisy-label image classification on CIFAR100-NL.
\end{abstract}}
\maketitle

\IEEEdisplaynontitleabstractindextext
\IEEEpeerreviewmaketitle

\begin{figure*}[th]
\centering
\setlength{\abovecaptionskip}{0.cm}
\includegraphics[scale=0.9]{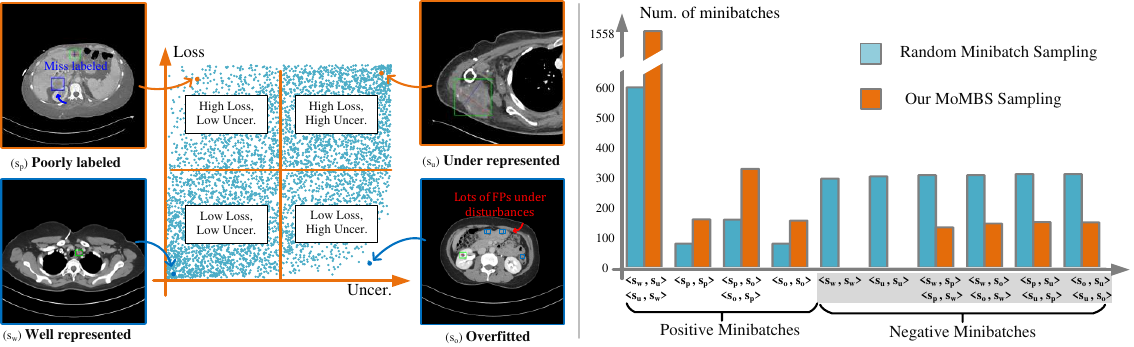}
\caption{\textbf{Left:} Training samples are typically grouped into four types based on data loss and uncertainty:  ($s_p$) poorly labeled, ($s_u$) under represented, ($s_w$) well represented and ($s_o$) overfitted samples. The loss-based sample quality measurer, employed in self-paced curriculum learning (SCL) and online hard example mining (OHEM), inaccurately treats both ($s_p$) and ($s_u$) as low-quality samples.  \textbf{Right:} The distribution comparison of different minibatches obtained with random vs. the proposed mixed-order minibatch sampling (MoMBS). The MoMBS first categorizes a minibatch into positive and negative based on the types of training samples it contains and then, through mixed-order sampling, it increases the number of positive minibatchs and decreases the number of negative minibatchs, thereby enhancing deep network's parameters updating. In comparison, random MBS creates a large number of negative minibatchs. Note that the above plots and statistics are derived from the universal lesion detection experiment.}

\label{Fig1_overview}
\end{figure*}

\section{Introduction}\label{sec:intro}

\IEEEPARstart{D}{iverse-quality} images are commonly found in computer vision. In long-tailed image classification that exhibits a prevalence diversity (abundant vs. sparse), there is a high imbalance in the number of examples per class, thus forming under represented classes. In noisy-label image classification that exhibits label diversity (clean vs. noisy), there are images with labels that are manually or systematically corrupted. The hard-sample challenge becomes more pronounced in universal lesion detection (ULD) from computed tomography (CT), which focuses on localizing lesions of various types, rather than identifying the specific lesion categories. This is because ULD datasets often contain spotty images with lesions of diverse shapes and sizes. Consequently, this can lead to both the poorly labeled issue, such as mislabeling, incorrect labeling, and imprecise labeling, and the under represented issue, including blur, minority-class representation, tiny-lesion depiction, and confusion or overlap between different classes \cite{liu2020self} (see Fig. \ref{Fig1_overview}).

How to tackle diverse-quality training images is a significant concern in deep-learning-based computer vision tasks \cite{liu2020self}. There is a straightforward way of grouping hard images into two primary categories: \textit{poorly labeled} and \textit{under represented}. A \textit{poorly labeled} image is generally due to the labeling process, which can lead to erroneous or imprecise labels. For instance, images with semantically identical content may be annotated with differing labels. Conversely, an \textit{under represented} image predominantly emerges during the data acquisition process that yields blurry images or low prevalence of minority classes, which impedes the network's ability to effectively learn relevant information and thus leads to an under represented scenario.  An effective approach should aim to minimize the negative impact of poorly labeled images and prevent overfitting to incorrect labels, while simultaneously maximizing the utility of under represented samples to enhance the model's accuracy and robustness.

To address these issues, online sample-reweighting methods have been proposed to identify all high-loss samples as hard ones and adjust their importance during training. For instance, self-paced curriculum learning (SCL)~\cite{wang2021survey} dynamically evaluates the difficulty of individual samples based on their loss values, and subsequently de-emphasizes them in backward processes in a hard manner \cite{kumar2010self,liu2020self,liu2021co,ju2022improving} or a soft manners~\cite{jiang2014easy, zhao2015self,xu2015multi,gong2018decomposition}. In contrast, online hard example mining (OHEM) identifies hard samples based on their loss values and increases their significance by increasing the number of hard samples in subsequent training.
Despite the significant progress of sample-reweighting methods in natural image analysis tasks~\cite{jiang2014easy, zhao2015self,xu2015multi,gong2018decomposition}, they still confront two major challenges in addressing taskes like ULD. 

{\bf Firstly, the reliability of the loss-based measure for sample difficulty is questionable.} Although the deep network's loss value may reflect the sample difficulty to a certain degree, it is sometimes unreliable due to the network's confirmation bias\cite{Warburg_2021_ICCV,Mao_2021_ICCV}.  This issue is particularly notable in CT annotation, which is laborious and costly, easily leading to inconsistent annotations among different experts or sites. Since deep networks possess the capability to fit all samples, it may also fit noisy labels leading to a biased loss. Sample reweighting based on a biased loss can lead to \underline{under-utilization} or \underline{over-utilization} of these samples.
{\bf Secondly, the loss-based measure alone is insufficient to differentiate between poorly labeled and under represented samples}, even if we assume it  is reliable. For example, SCL may categorize the lesions from minority classes and lesions with wrong labels as hard samples, subsequently deweighting their losses as training proceeds. However, the minority classes of lesions are actually useful for improving network performance, thus deweighting their losses leads to sample under-utilization. Similarly, oversampling the incorrectly labeled images may have a negative impact on network training, leading to over-utilization of these samples.

To address the issues raised by hard training samples, our \underline{first contribution} of this paper lies in better characterizing training samples using both loss and uncertainty metrics, instead of using loss only. As shown in  Fig. \ref{Fig1_overview}(Left), based on data loss $l$ and uncertainty $u$, four distinct data categories emerge:
\begin{enumerate}
    \item[\textbf{($s_p$)}] {Data with a high loss $l^h$ and a low uncertainty $u^l$, likely indicating \textbf{poorly labeled samples} that are mislabeled or wrongly-labeled};
  \item[\textbf{($s_u$)}] {Data with a high loss $l^h$ and a high uncertainty $u^h$, likely representing \textbf{under represented samples} that have insufficient samples or are in conflict with majority-class};
  \item[\textbf{($s_w$)}] {Data with a low loss $l^l$ and a low uncertainty $u^l$, likely corresponding to \textbf{well represented} samples that are well-learned by the network or those from majority-class samples};
  \item[\textbf{($s_o$)}] {Data with a low loss $l^l$ and a high uncertainty $u^h$, likely indicating  \textbf{overfitted samples}, from which the network learns to fit wrong information}.
\end{enumerate}



Our \underline{second contribution} involves introducing a novel minibatch sampling (MBS) approach to effectively handle the above issues. We argue that \textbf{minibatch sampling plays a critical role in addressing the challenges posed by hard training data}. Hence, as shown in Fig. \ref{Fig1_overview}(Right), we categorize a minibatch into positive and negative intuitively based on the four distinct data categories. Prior to us, only a few studies have suggested that training samples within the same minibatch influence each other's training, thereby affecting the overall  performance~\cite{dokuz2021mini}. However, these studies lack comprehensiveness in both theoretical analysis and experimental validation. Furthermore, there is a lack of insight into how to design an effective MBS method to tackle these issues.

Our \underline{last contribution} is we further provide a experimental explanations of our minibatch categorization from a novel perspective, update efficacy, besides the intuitive explanation in the second contribution. A positive minibatch triggers a reasonable update to the network parameters while a negative minibatch brings a low effective network parameters update.


Our proposed MBS approach is called  \textbf{mixed-order minibatch sampling (MoMBS)}. MoMBS is designed to increase the number of positive minibatches and significantly reduce the number of negative minibatches, thereby enhancing the utilization of training samples based on their data category.
For example, we construct a minibatch that consists of well represented samples and under represented samples, instead of combining poorly labeled samples with poorly labeled or overfitted samples. By doing so, the under represented samples can be the primary contributor to gradient calculation in its iteration, while poorly labeled or overfitted samples exert a less influence on other training samples.

MoMBS consists of an assessor and a schedule. 
{\it The assessor} calculates the loss and uncertainty for each sample, ranks the samples based on each measure, and computes the sum of the rank indices to represent each sample's difficulty.  We use the sum of rank indices rather than the sum of loss and uncertainty values to address the limitations of fluctuations in network training and the lack of comparability between the scales of loss and uncertainty.  {\it The scheduler} simulates human perception behavior to sample a minibatch. Humans can
easily lose concentration and fail to learn if all samples are of the same difficulty. Therefore, we argue that during network training, the samples in a minibatch should be mixed in terms of their difficulties and MoMBS follows this human perception behavior. Specifically, MoMBS aims to maintain consistent total difficulties (i.e., the sum of loss index and uncertainty index) for each minibatch during a training epoch. To achieve this, samples of high difficulty are paired with those of low difficulty. As elaborated further on, this straightforward approach increases the number of positive minibatches and significantly minimizes the number of negative minibatches, thereby optimizing the use of training samples according to their data category. It is worth noting that even when the estimated loss and uncertainty are not reliable with respect to the ground truth sample difficulty, our MoMBS has a minor negative effect on network training because no sample reweighting is used.

Obviously, the loss and uncertainty can be unreliable in some tasks with training samples of extremely diverse quality like long-tailed (LT) and noisy-label (NL) image classification. However, our experiments show that the training samples still adhere to the proposed categorization to some extent; therefore, MoMBS can also be helpful in these tasks. We evaluate the effectiveness of our MoMBS on ULD task based on two state-of-the-art (SOTA) ULD methods, validate the generalization ability on long-tailed (LT) image classification task, noisy-label (NL) image classification task, and COVID CT segmentation task. Our extensive experiments demonstrate that  \textbf{MoBMS consistently improves the performance of all four tasks without requiring extra special network designs}.

\section{Related Work}

\subsection{Self-paced Curriculum Learning (SCL)}
SCL is a type of curriculum learning (CL) method \cite{wang2021survey,liu2022acpl,morerio2017curriculum,binkowski2019batch,kong2021adaptive,wang2019dynamic}, in which the sample difficulty measure and the training scheduler are both designed in a data-driven manner. Specifically, SCL evaluates a sample's difficulty based on its loss value and reduces the weight of losses associated with hard samples during subsequent training phases. Kumar {\it et al.} \cite{kumar2010self} introduce the concept of SCL to deactivate the highly-difficult samples by incorporating a hard self-paced regularizer (SP regularizer). The early attempt of SCL inspires the studies of new SP regularizers to enhance the utilization of different samples in network training. These regularizers include linear \cite{jiang2014easy,liu2020self}, logarithmic \cite{jiang2014easy}, mixture \cite{jiang2014easy,zhao2015self}, logistic \cite{xu2015multi}, and polynomial \cite{gong2018decomposition}  variations. Despite the effectiveness of such methods, such a loss-based sample deweighting mechanism can unavoidably cause the sample under-utilization issue. Furthermore, extensive efforts have been invested in exploring the theoretical underpinnings of SCL \cite{meng2017theoretical} , yielding wide visual category
discovery \cite{lee2011learning}, image segmentation \cite{kumar2011learning,yang2022micl}, image
classification \cite{tang2012self,liu2021co,ju2022improving}, object detection \cite{tang2012shifting,zhang2019leveraging}, object retrieval \cite{jiang2014easy}, person re-identification (ReID) \cite{zhou2018deep}, etc. SCL verifies the usefulness of pseudo label generation\cite{jiang2014easy,han2019weakly,ghasedi2019balanced} during model training. Researchers also adopt group-wise weight based on SCL, {\it e.g.}, multi-modal \cite{gong2016multi}, multi-view \cite{xu2015multi}, multi-instance \cite{zhang2015self}, multi-task \cite{li2017self2}, etc. Additionally,
SCL has found application in data-selection-based training strategies, {\it e.g.}, active learning \cite{lin2017active,tang2019self}.

\subsection{Uncertainty Estimation}
Existing uncertainty estimation techniques can be classified into two categories: Bayesian and non-Bayesian methods.
Bayesian methods model a neural network's parameters as a posterior distribution using input data samples to derive the probability distributions for output labels~\cite{mackay1992practical}.
Given the intractability of this posterior distribution, some approximate variants of Bayesian modeling have been proposed for Bayesian methods, {\it e.g.}, Monte Carlo dropout~\cite{gal2016dropout} and Monte Carlo batch normalization~\cite{teye2018bayesian}. Non-Bayesian methods like Deep Ensembles~\cite{lakshminarayanan2016simple}  train multiple models and employ their variance to quantify the uncertainty. Moreover, uncertainty estimation techniques  \cite{ju2022improving,9447177pmi,9864260pmi,9699392pmi,10173747pmi,10079153pmi,10158031pmi,10223424pmi,9864260pmi,9801711pmi,10302334pmi,9086445pmi,9745778pmi} have been used to enhance the analysis of medical images. In this work, we use uncertainty as a measure of the sample's quality.

\subsection{Online Hard Example Mining (OHEM)}
OHEM~\cite{shrivastava2016training,Dong_2017_ICCV,Sun_2019_ICCV,he2021online,8733121pmi} is widely used in various tasks such as image segmentation and object detection. The core idea involves dynamically selecting hard samples ({\it e.g.}, triggering a high loss) and oversampling them during network training. While OHEM has achieved success, 
it can easily introduce wrong information when the training data contains lots of samples with inaccurate or wrong labels.

\subsection{Long-tailed (LT) Image Classification}
Concerning the LT issue \cite{10105457pmi,9848833pmi,10192363pmi,9774921pmi,10105457pmi,10124024pmi,9858006pmi,10238823pmi}, there are three main directions to improve the  classification performance: i) Loss modification, including sample-wise re-weighting methods \cite{lin2017focal,ren2018learning} and Class-wise re-weighting methods \cite{cui2019class,huang2016learning,khan2017cost,tan2020equalization,park2021influence,li2022long}; ii) Logit adjustment, which assigns relatively large margins for tail
classes \cite{menon2020long,hong2021disentangling,tang2020long,zhang2021distribution,cao2020domain,cao2019learning}; and iii) Decoupling representation, which focuses on improving the LT performance by decoupling the representation and classifier \cite{wang2020devil,zhang2021distribution,zhong2021improving,zhou2020bbn,wang2021contrastive}.
None of them considers the aspect of MBS, hence we can apply our MoMBS to some of them without any conflict.

\subsection{Noisy-label (NL) Image Classification}
The existing works on deep learning with noisy labels can be classified into five categories by exploring different strategies \cite{song2022learning}:  i) Robust architectures, which add a noise adaptation layer at the top of an underlying deep learning network (DNN) to learn a label transition process or developing a dedicated architecture to reliably support more diverse types of label noise \cite{xiao2015learning,chen2015webly,goldberger2016training,han2018masking,cheng2020weakly,jindal2016learning,goldberger2016training,lee2019robust,10039708pmi,9790332pmi,9674931pmi,9784878pmi,9477020pmi,10209198pmi}; ii) Robust regularization that enforces a DNN to overfit less to false-labeled examples explicitly or implicitly \cite{tanno2019learning,menon2019can,xia2020robust,wei2021open}; iii) Robust loss function designs to improve the loss function \cite{pereyra2017regularizing,lukasik2020does}; iv) Loss adjustment that adjusts the loss value according to the confidence of a given loss (or label) by loss correction, loss reweighting, or label refurbishment \cite{wang2019symmetric,feng2021can,liu2020peer,amid2019robust,ma2020normalized,arazo2019unsupervised,yao2020dual,liu2015classification,zhang2021dualgraph,zheng2020error,chen2021beyond,shu2019meta,zhang2020distilling,li2017learning,zheng2021meta} ; and v) Sample selection: identifying true-labeled examples from noisy training data via multi-network or multi-round learning \cite{song2019selfie,chen2019understanding,song2021robust,nguyen2019self,li2019dividemix,wei2020combating,huang2019o2u,wu2020topological,wu2021ngc,zhou2020robust}.  None of them considers the aspect of MBS, hence our proposed MoMBS works seamlessly with them.

\subsection{COVID CT Segmentation}
Since December 2019, a novel Coronavirus Disease (COVID-19) has caused a global health crisis to the world. COVID-19 lesion segmentation \cite{10032636,9969651,9931157,9810271,cao2020longitudinal,huang2020serial,wang2020noise,fan2020inf,9115057,wang2020weakly,wu2021jcs,liu2021covid,yao2021label} is an active area and helps ease the burden for radiologists.While achieving success, the heterogeneity of COVID-19 lesions remains a challenge that hinders their performance. However, all the above methods use a standard MBS strategy.

\begin{figure*}
\centering
\includegraphics[scale=0.77]{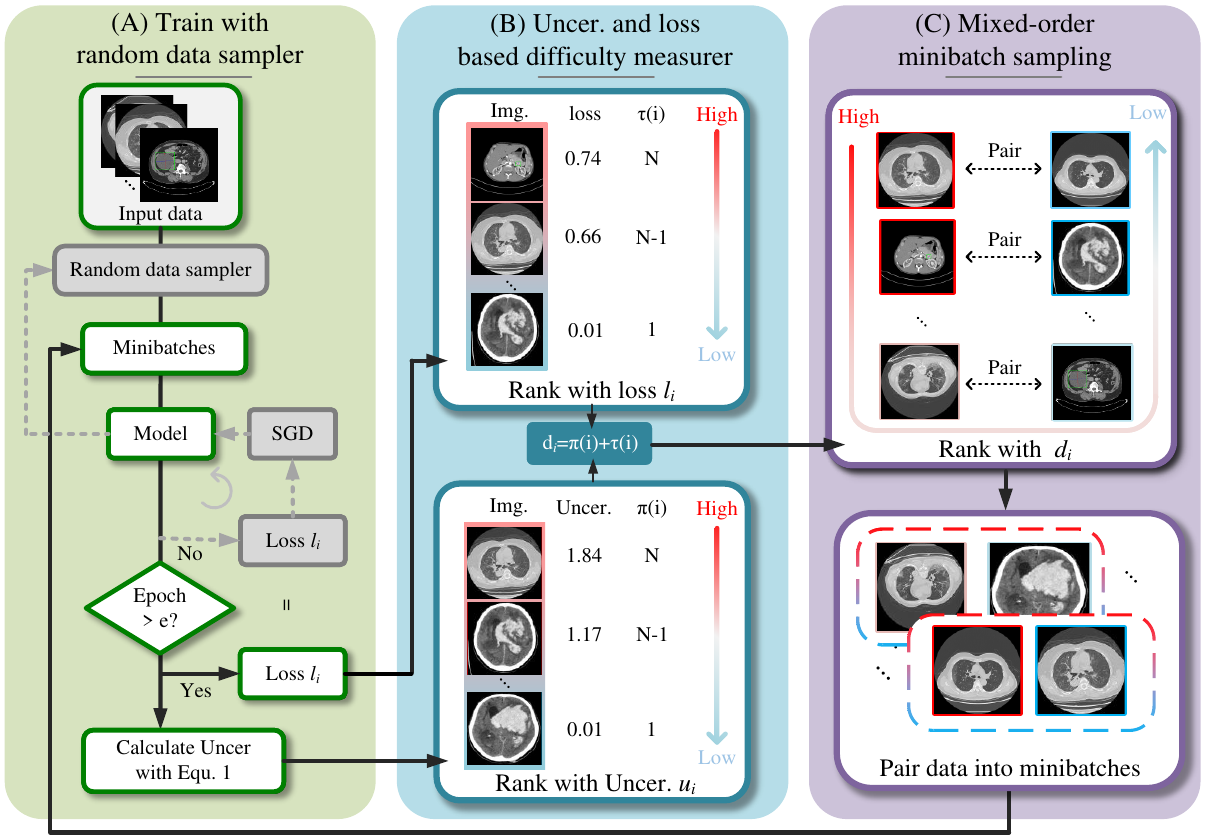}
\caption{Our MoMBS method consists of three learning steps. In step A, we randomly sample data using the vanilla random sampler for initial  network training. After $e$ epochs, we activate the uncertainty estimation component (Eq. \ref{Equ1}) to calculate the uncertainty of each sample. In step B, we sort all samples based on their loss and uncertainty ranks and calculate the sum of the rank indices of uncertainty $\pi(i)$ and rank indices of loss $\tau(i)$ to obtain their difficulty rank score $d$. Finally, in step C, we rank all training samples based on their difficulty rank score $d$ and construct minibatches by pairing samples with high $d$ with those with low $d$. The newly formed minibatches are used for the subsequent network training.}  \label{fig2_network_architecture}
\end{figure*}

\subsection{Universal lesion Detection (ULD)}
Computed tomography (CT)-based ULD, serves as a crucial component in computer-aided diagnosis (CAD)  by localizing diverse lesion types. Despite its clinical significance, ULD is fraught with challenges due to the heterogeneity of lesion shapes and types, and the resource-intensive annotation process. Most existing ULD methods incorporate several adjacent 2D CT slices 
as the 3D context information for 2D detection network  \cite{zhang2020Agg_Fas,yan20183DCE,li2019mvp,yan2019mulan,yang2020alignshift,zhang2020revisiting,tang2021weakly,yang2021asymmetric,li2021conditional,lyu2021segmentation} or directly adopt 3D designs \cite{cai2020deep}.

\section{Mixed-order minibatch sampling (MoMBS)}
This section provides a detailed description of our MoMBS, including the problem definition in Section \ref{sec.problem}, the sample difficulty assessor in Section \ref{sec.measurer}, and the minibatch sampling scheduler in Section \ref{sec.motrain}. We also provide explanations for our proposed minibatch categorization in Section \ref{sec.theory}.
\subsection{Problem Definition} \label{sec.problem}
The training dataset is represented as  $\mathcal{X}=\{(x_i,y_i)\}_{i=1}^{I}$, where $x_i$ denotes the $i$-th input data and $y_i$ denotes the corresponding label. The primary objective of the task can be generally represented as  $\hat{y_i}=\mathcal{F}(x_i|w)$ with parameter $w$ to establish a mapping from $x_i$ to  $y_i$. The difference between  $\hat{y_i}$ and  $y_i$ is measured by a risk function (or loss function). We adopt the widely-used stochastic gradient descent (SGD) for risk function minimization. The vanilla SGD iteratively updates the model weight $w$ based on a minibatch $\mathcal{B}$, which is sampled from the training data according to a certain strategy.

\subsection{Sample Difficulty Assessor} \label{sec.measurer}
We derive the sample difficulty measurement following three steps: (i) uncertainty estimation, (ii) loss- and uncertainty-based ranking, and (iii) sample difficulty score computation. The derived difficulty score is used for the subsequent minibatch categorization.

\textbf{(i) Uncertainty estimation.} As depicted in Step A of Fig.~\ref{fig2_network_architecture}, the uncertainty estimation process is initiated after a certain pivot epoch $e$. For an image $x_i$ and the current model $\mathcal{F}(\cdot|w)$, we calculate the uncertainty $u_i$ as the information entropy of the model's average prediction $\hat{y_i}=\mathcal{F}(x_i|w)$ under $G$ disturbances ${t^{1},t^{2},...,t^{G}}$:

\begin{equation}
\begin{aligned}
   \hat{y}_{i}^{g}=\mathcal{F}(x_i|w, t^{g} ),~~
   u_i=H(\frac{1}{|G|}\sum_{g=1}^{G} \hat{y}_{i}^{g}),
 \end{aligned} \label{Equ1}
  \end{equation}
where $H$ is the information entropy, and $\hat{y}_{i}^{g}$ represents the prediction of $\mathcal{F}(x_i)$ under disturbance $t^{g}$. It is worth noting that we directly introduce noise $t^{g}$ into key feature maps ({\it e.g.,} the output of the encoder in the segmentation network, or the backbone feature maps in two-stage detection methods) rather than into the input images. This is because adding noise to input images does not substantially alter the network's output ~\cite{ouali2020semi}. The feature map $\hat{f}_{g}$ under a disturbance $t^{g}$ is formulated as follows:
    \begin{equation}
      \begin{aligned}
          \hat{f}_{g}= f \otimes (\mathds{1}+t^{g}),
      \end{aligned}
    \end{equation}
   where $\otimes$ is the pixel-level dot multiplication, $\mathds{1}$ is a matrix with the same size as \textit{f} and filled with $1$.
   Each pixel in $t^{g}$ is sampled from a uniform distribution $U[-\gamma,+\gamma]$.

\textbf{(ii) Loss- \&  uncertainty-based ranking.} As shown in Fig. \ref{fig2_network_architecture} (step B), we rank all training samples in a descending order according to their respective loss and uncertainty values.
\begin{equation}
  \begin{aligned}
      \mathcal{X}_u= \{(x_{\pi(j)}, y_{\pi(j)})\},~~~~s.t.~~u_{\pi(j)}\geq u_{\pi(j+1)}; \\
      \mathcal{X}_l= \{(x_{\tau(k)}, y_{\tau(k)})\},~~~~s.t.~~l_{\tau(k)}\geq l_{\tau(k+1)},
  \end{aligned}
\end{equation}
where $\pi(j)$ and $\tau(k)$ is the indices for the $j$-th and $k$-th ranked sample in terms of uncertainty $u$ and loss $l$ values, respectively.
This ranking approach overcomes the challenges of training fluctuations and the incomparable value scales between loss and uncertainty.

As shown in  Fig. \ref{Fig1_overview}, based on the ranked data loss and uncertainty values, four distinct data scenarios emerge to form a \underline{sample categorization}:
 \begin{enumerate}
 \item[\textbf{($s_p$)}] {Data with a high loss $l^h$ and a low uncertainty $u^l$ suggests that while the prediction is inconsistent with the labels, the network has a high confidence in its prediction. This could indicate that the data classes are mislabeled or incorrectly labeled, representing \textbf{poorly labeled} samples};
 \item[\textbf{($s_u$)}] {Data with a high loss $l^h$ and a high uncertainty $u^h$ signifies that the prediction is inconsistent with the labels, and the prediction can be significantly influenced by disturbances. This might indicate that the data is under represented by the network, possibly due to insufficient samples for their classes or conflicts with majority-class data. These are \textbf{under represented} samples};
 \item[\textbf{($s_w$)}] {Data with a low loss $l^l$ and a low uncertainty $u^l$ indicates that the prediction is consistent with the labels, and the network is confident with its prediction. This could correspond to well-learned or majority-class samples, which are \textbf{well represented} samples};
 \item[\textbf{($s_o$)}] {Data with a low loss $l^l$ and a high uncertainty $u^h$ suggests that the prediction aligns with the label but can be significantly influenced by disturbances. This represents \textbf{overfitted} samples}.
 \end{enumerate}

\textbf{(iii) Difficulty score computation.}
Similar to the loss-based difficulty assessor in OHEM and SCL, the above sample categorization faces occasional unreliability. Additionally, the levels of loss and uncertainty associated with the samples are not simply categorized as high or low; in fact, a majority of them fall into the medium category. As a result, directly reweighting the sample based on these categorization results as in OHEM and SCL would be suboptimal.

In our method, we leverage under represented samples $s_u$ by pairing them with well represented samples $s_w$, rather than directly decreasing the number of negative samples, namely poorly labeled $s_p$ and overfitted $s_o$ samples, in situations where the difficulty assessor is less reliable. We only need to distinctly categorize under represented samples $s_u$ and well represented samples $s_w$. Therefore, we directly sum the indices in $\mathcal{X}_u$ (i.e., $\pi(i)$) and $\mathcal{X}_l$ (i.e., $\tau(i)$) to obtain a difficulty rank score $d$ for each training sample:
\begin{equation}
  \begin{aligned}
    d_{i}=\pi(i) + \tau(i).\\
  \end{aligned}
\end{equation}
The low (or high) difficulty rank score $d$ indicates both the loss and uncertainty of the sample are low or high. This approach enhances the robustness and efficacy of the method.
In general, a well represented sample $s_w$
has a low difficulty score $d$, a under represented sample $s_u$ has a high difficulty score $d$, and a poorly labeled sample $s_p$ or an overfitted sample $s_o$ has a medium difficulty score $d$. That is,
\begin{equation}
  d(s_w)~<~ d(s_p) ~or~ d(s_o) ~<~d(s_u).
\end{equation}

\subsection{MoMBS Scheduler} \label{sec.motrain}
We first describe different minibatch sampling strategies: random minibatch sampling (RaMBS), SCL, OHEM, and our MoMBS. Then, we illuminate the differences in the minibatchs produced by them and subsequently analyze the impact of these varied minibatch productions. To simplify our explanation, we set the minibatch size $b$ to $b=2$.

\subsubsection*{\bf Minibatch formulation}

In \underline{Random MBS}, two samples are randomly selected without replacement from the entire training dataset $\mathcal{X}=\{(x_i,y_i)\}_{i=1}^{I}$ to compose a minibatch $B_{\cdot}$ in each training iteration:
\begin{equation}
 \begin{aligned}
&B_{\cdot}= <x_m,x_n>,~~~s.t.~~m \neq n,~~ m,n\in \{1,2,\ldots,N\}; \\
&B_{i} \cap B_{j} = \varnothing,~~~s.t.~~i \neq j,~~ i,j\in \{1,2,\ldots,N/2\}.
 \end{aligned}
\end{equation}
This process is performed repeatedly until all training samples have been sampled to complete an epoch.

As for \underline{SCL and OHEM}, the random MBS mechanism is still utilized, but the occurrence or importance of certain samples in the entire training dataset $\mathcal{X}$ is modified. Specifically, SCL identifies hard samples and decreases their occurrence in $\mathcal{X}$ or their weight in the loss calculation. Conversely, OHEM increases the occurrence of hard samples in $\mathcal{X}$ or their loss weight. As a result, OHEM and SCL can face the issues of sample under- or over-utilization.

In \underline{proposed MoMBS}, we introduce a novel method of constructing minibatches without altering the occurrence or importance of the training samples, thereby avoiding the issue of sample under-or over-utilization. Inspired by the observation that humans tend to be more attentive when presented with a mixture of easy and challenging tasks, we pair samples with a high difficulty rank score $d$ with those with a low $d$ within a minibatch. This is done with the aim of keeping the total difficulty score of samples evenly distributed as much as possible across all minibatches, as illustrated in Step C of Fig. \ref{fig2_network_architecture}. Formally,
\begin{equation}
 \begin{aligned}
&d(B_{i})=d_m+d_n, ~&s.t.~B_{i}= <x_m,x_n>, \\
&{\cal B}= \arg\min_{\cal B}~Var(d({\cal B})) ~&s.t.~{\cal B}=\{B_1,\ldots,B_{N/2}\},
 \end{aligned}
\end{equation}
where $d(B_{i})$ represents the total sample difficulty score of a minibatch $B_{i}$, $d({\cal B})$ represents the set $d({\cal B})=\{d(B_1),\ldots,d(B_{N/2})\}$, and $Var(.)$ computes the variance.

\subsubsection*{\bf Categorization of minibatches produced by MoMBS}

As mentioned above, we categorize training samples into four distinct types: poorly labeled ($s_p$), under represented ($s_u$), well represented ($s_w$), and overfitted ($s_o$). Consequently, this results in ten possible minibatch types as a minibatch $<s_{\cdot}, s_{\cdot}>$ contains two samples and the order of the two samples with a minibatch does not matter. These ten types are further categorized into two classes, depending on whether a minibatch is effective or not for network training (the categorization reasons will be explained later):

%
%
%
%
%

\begin{itemize}
    \item  Positive minibatches:

    MB$_1$~$<s_w, s_u>|<s_u, s_w>$;~~MB$_2$~$<s_p, s_p>$;

    MB$_3$~$<s_p, s_o~>|<s_o, s_p~>$;~~MB$_4$~$<s_o, s_o>$.
\end{itemize}


With this minibatch categorization, there are four distinct types of positive minibatches: MB$_1$ directs the network to focus on under represented samples $s_u$, while its impact on well represented samples $s_w$ is minimized due to their high robustness to the network updating. As for MB$_2$, MB$_3$ and MB$_4$, they group together hard samples, such as those that are poorly labeled $s_p$ or overfitted $s_o$. Employing this strategy mitigates the risk that these samples adversely affect other categories, particularly the under represented samples.

%
%
%

\begin{itemize}
\item Negative minibatches:

MB$_5$~$<s_w,s_w>$;~~MB$_6$~$<s_u,s_u>$;

MB$_7$~$<s_w,s_p>$ ~~~or~~~ $<s_p,s_w>$;

MB$_8$~$<s_w,s_o>$ ~~~or~~~ $<s_o,s_w>$;

MB$_9$~$<s_p,s_u>$ ~~~or~~~ $<s_u,s_p>$;

MB$_{10}$~$<s_o,s_u>$~~~or~~~$<s_u,s_o>.$
\end{itemize}

%

In section \ref{sec.theory}, we will show that a lower loss brings a less contribution to network update. Therefore, MB$_5$ typically has a minimal impact on network update due to the low loss of $s_w$. In the case of MB$_6$, achieving a mutually beneficial outcome is often challenging. The high magnitude of network update and the diminished robustness of network update make it a delicate balance. MB$_7$ or MB$_8$ tends to direct the network to overly emphasize poorly labeled samples $s_p$ or overfitted samples $s_o$. Consequently, this can degrade network performance due to reliance on inaccurate labels or can accentuate the overfitting problem. MB$_9$ and MB$_{10}$ often struggle to enhance the network's learning from under represented samples $s_u$ due to the high loss (MB$_9$) and low robustness to the network updating (MB$_{10}$). Moreover, the presence of these minibatch types diminishes the probability of MB$_1$'s occurrence.

As depicted in Fig. \ref{Fig1_overview}, a random sampling mechanism can yield numerous negative minibatches. In contrast, our MoMBS approach significantly reduces, or even eliminates, these negative minibatches, while increasing the number of positive minibatches. This is achieved because MoMBS maintains a consistent total difficulty score across all minibatches throughout the entire training dataset, thereby significantly reducing the probability of certain combinations (e.g., the total rank score of $<s_w, s_w>$ is too low ), while increasing the probability of others (e.g., the total rank score of $<s_w, s_u>$ is optimal).

\newsavebox{\tablebox}

\begin{figure}[t]
\centering
\setlength{\abovecaptionskip}{0.cm}
\includegraphics[scale=0.52]{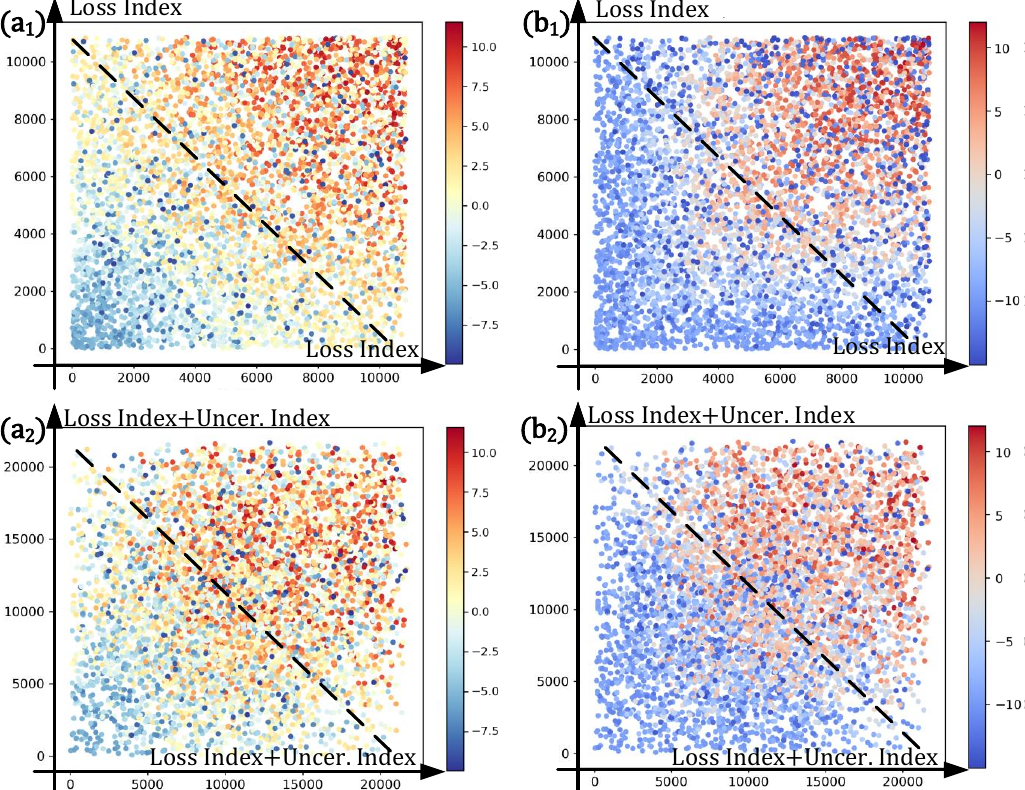}
\caption{($a_{1}$): Total loss reduction after one iteration backward for two samples in a minibatch vs. their individual loss values. ($b_{1}$): Loss reduction after one iteration backward of the sample with the lower loss in a minibatch vs. their individual loss values. ($a_{2}$): Total loss reduction after one iteration backward vs. the sum of their individual loss and uncertainty values.  ($b_{2}$): Loss reduction after one iteration backward of the sample with the lower loss in a minibatch vs. the sum of their individual loss and uncertainty values.}
\label{update_explain}
\end{figure}

\subsection{Explanations of MB categorization}\label{sec.theory}

Contrasting with the intuitive explanation based on categorizing four training sample types using loss and uncertainty, we now provide a experimental explanation of our minibatch categorization from a novel perspective\textemdash update efficacy. Update efficacy evaluates the actual effectiveness of each training iteration by measuring the extent to which network parameter adjustments contribute to model convergence and performance.
In this section, we first demonstrate that, despite its limitations, loss can act as an updated efficacy measure. Subsequently, we show how its integration with uncertainty can partially mitigate the limitations.

\subsubsection*{\bf Using loss to measure update efficacy}
In this section, we aim to show that loss can effectively measure update efficacy.
Our proof is based on a sigmoid-based (or softmax-based) network ${\hat y}_i=\mathcal{F}(x_i|w_t)=\sigma(x_i^{T}w_t)=\sigma(z_i)$, where $w_t$ is the network weight at iteration $t$, $\sigma$ is the sigmoid function for binary classification tasks (or softmax for multi-class classification tasks), and $z$ is the latent feature input of the sigmoid (or softmax) function. For illustration, we use the Cross-Entropy (CE) loss for a minibatch $B=<x_1,x_2>$:
\begin{equation}
  \begin{aligned}
        l(B)=\frac{1}{2}\sum_{i=1}^{2} l_{i},
~l_{i} = - y_{i} \log(\hat{y}_{i}) - (1-y_{i}) \log(1-\hat{y}_{i}).
  \end{aligned}
\end{equation}
We assume that $w_t$ represents the reasonably converged weight with a random sampling manner over the whole training dataset. Then we use the SGD to optimize $w$:
\begin{equation}
 w_{t+1}= w_t - \eta g_t,~~~ g_t=\frac{ \partial l(B)}{\partial w_t}=\frac{1}{I}\sum_{i=1}^{I} \frac{ \partial l_{i} }{ \partial w_t},
\label{Equ10}
\end{equation}
where $\eta$ is the learning rate, $g_t$ is the gradient at time $t$. Without simple mathematical derivation, it can be shown that
\begin{equation}
 \begin{aligned}
    \frac{\partial l}{\partial \hat{y}_{i}} &= \frac{y_{i}}{\hat{y}_{i}} - \frac{1-y_{i}}{1-\hat{y}_{i}}, ~\quad  \frac{\partial \hat{y}_{i}}{\partial z_{i}} = \hat{y}_{i} (1-\hat{y}_{i}),\\
    \frac{ \partial l_{i} }{ \partial w_t}&=\frac{ \partial l_{i} }{ \partial \hat{y}}\frac{ \partial \hat{y}_i }{ \partial z_i} \frac{ \partial z_i}{ \partial w_t}=(\hat{y}_{i} - y_{i}) \frac{ \partial z_i}{ \partial w_t}.
 \end{aligned}
\label{Equ15}
\end{equation}
From the above, it becomes evident that the gradient of the loss is influenced by two main components: the prediction error $(\hat{y}_{i} - y_{i})$ and the input data (or feature maps). The first factor often holds more influence due to: i) Normalization impact. Deep networks typically use the layers such as Batch Normalization (BN) to normalize activations and gradients. This process can mute variations from the feature map, highlighting the prediction error's role in gradient updates; and ii) Chain rule sensitivity in back-propagation, that is, factors closer to the output have more influence on the gradient in deep networks. This is because their impact spans from the last layer of the network to the first.

Given the positive correlation between prediction error $ \hat{y}_{i} - y_{i} $ and loss, we initially deduce that a minibatch $B$ with a higher loss value should exhibit greater update efficiency. We hereby measure the update efficiency of one minibatch via its loss reductions $\Delta l_B$ after optimizing the model based on their loss.
\begin{equation}
 \Delta l_B= l(B|\mathcal{F}(\cdot|w_t))-l(B|\mathcal{F}(\cdot|w_{t+1})).
\label{Equ_update_efficiency}
\end{equation}
As depicted in Fig. \ref{update_explain} (a1), we demonstrate each minibatch's loss reductions (colors) vs. the loss values of the two samples in the minibatch (x and y axis). It is observed that a higher total loss of a minibatch (top right) results in a more substantial loss reduction, whereas a lower total loss leads to a lesser or even negative reduction. Ideally, filtering out minibatch combinations with low or negative update efficacy would be optimal, but this is not feasible due to the necessity of traversing all training samples. Hence, maintaining an even total loss value across all minibatches (i.e., diagonal from top left to bottom right) emerges as a practical approach.

However, a per-sample analysis exposes some limitations of this method. Now recall the loss gradient calculation of a minibatch $B$ in Equ. \ref{Equ15}. Observations also suggest that samples with a larger loss in a minibatch have a greater impact on the total gradient. Therefore, our solution, which involves selecting diagonal minibatches from top left to bottom right, should also ensure it does not result in too low or even negative update efficacy, particularly for samples $x_{min}$ with lower loss values.
\begin{equation}
  \begin{aligned}
x_{min} &=
\begin{cases}
x_1 & l_1 < l_2, \\
x_2 & else.
\end{cases}\\
\Delta l_{x_{min}}&= l(x_{min}|\mathcal{F}(\cdot|w_t))-l(x_{min}|\mathcal{F}(\cdot|w_{t+1})).
\end{aligned}
\end{equation}
Fig. \ref{update_explain} (b1) illustrates the loss reduction for  $x_{min}$. We can find that maintaining an even total loss value across all minibatches (dotted line) can still lead to low or negative update efficacy, particularly those in the left top or bottom right. To address this issue, our work incorporates uncertainty as an additional factor.

\subsubsection*{\bf Use uncertainty to measure the robustness}

Uncertainty is commonly used to assess a network's reliability or robustness. Ideally, we should calculate uncertainty across various network updates, ${\Delta w^{1},...,\Delta w^{G}}$. Yet, identifying appropriate disturbances for network parameters is challenging due to varying magnitudes across layers and the requirement of careful designs. In our method, we introduce disturbances $t^{g}$ into crucial feature maps $f$ (e.g., encoder output in segmentation networks, or backbone feature maps in two-stage detection methods) to mimic changes in network parameters as in Equ. \ref{Equ1}. The derived uncertainty signifies the network's robustness for sample $x_i$, which differs from the role played by loss. We simply sum the uncertainty and loss as a difficulty score $\hat {d}$ of the training samples:
\begin{equation}
  \begin{aligned}
    \hat {d_{i}}=u_i + l_i.\\
  \end{aligned}
\end{equation}
Fig. \ref{update_explain} (a2) shows that the difficulty score shows a similar trend with to loss value. However, maintaining an even total difficulty score $\hat {d}$ across all minibatches, alleviates the limitations in our loss-only methods.


\subsubsection*{\bf Consistency between experimental and intuitive MB categorization}
In our experimental proof of minibatch categorization, we denote minibatches along the diagonal from top left to bottom right as positive, and others as negative. This is consistent with our intuitive minibatch categorization that initially classifies training samples into four categories based on loss and uncertainty and then analyzes the ten possible minibatch combinations arising from these categories.

\begin{table*}[!h]
\centering
\caption{Sensitivity ($\%$) at various FPPI on the official testing dataset of DeepLesion \cite{yan18deeplesion} (upper) under 25$\%$, 50 $\%$ and 100 $\%$ training data settings with batchsize = 4, or on revised \cite{cai2020lesion}  testing dataset of DeepLesion \cite{yan18deeplesion} (lower) under 25$\%$ and 50 $\%$ training data settings with batchsize = 4. SCL and OHEM denote the self-paced curriculum learning and online hard example mining, respectively.  Mo+l, Mo+u, Mo+l+u denote using loss-base, uncertainty-based, and loss+uncertainty-combined  difficulty measurers with our Mixed-order scheduler. $\hat{Mo}$ denotes anti-mixed-order data pairing which pairs high- (or low-) difficulty data with high- (or low-) difficulty data into one minibatch, i.e., hi+hi.} \label{results}

\begin{lrbox}{\tablebox}
\begin{tabular}{p{28mm}p{15mm}p{13mm}p{13mm}p{14mm}p{15mm}p{23mm}p{23mm}p{23mm}p{23mm}}

  \toprule[1.5pt]
  &&&&&\textbf{ORIGINAL}&\textbf{TESTSET}&&&\\
  \midrule
  \textbf{Methods}&\textbf{Category}&\textbf{Data}&\textbf{Measurer}&\textbf{Sample}&\textbf{Deweight}&\textbf{$@0.5$}&\textbf{$@1$}&\textbf{$@2$}&Avg.[0.5,1,2]\\
\midrule
A3D\cite{yang2021asymmetric}&Baseline&25\% &-& Random &-&55.67  &65.39 &73.35& 64.80 \\

A3D+deweight\cite{jiang2014easy}&SCL\cite{jiang2014easy}&25\% &loss& Random &Hard Liner&54.28 (1.39$\downarrow$) &63.99 (1.40$\downarrow$) &72.18 (1.17$\downarrow$) &63.48 (1.32$\downarrow$) \\

A3D+deweight\cite{liu2020self}&SCL\cite{liu2020self}&25\% &loss& Random &SPE&55.54 (0.13$\downarrow$) &65.51 (0.12$\uparrow$) &72.44 (0.91$\downarrow$)&64.50 (0.30$\downarrow$) \\

A3D+retrain\cite{Dong_2017_ICCV}&OHEM\cite{Dong_2017_ICCV}&25\% &loss& Random &$\times$2&56.14 (0.47$\uparrow$) &65.97 (0.58$\uparrow$) &72.66 (0.69$\downarrow$)&64.92 (0.12$\uparrow$) \\

A3D+$\hat{Mo}$+l+u&Ablation&25\% &loss+uncer.& hi+hi &-&56.90 (1.73$\uparrow$)&66.19 (1.80$\uparrow$) &74.63 (1.28$\uparrow$) &65.90 (1.10$\uparrow$)\\

A3D+Mo+l+u&Ours&25\% &loss+uncer.& hi+low &-&\textbf{60.34} (4.67$\uparrow$) &\textbf{69.38} (3.99$\uparrow$)&\textbf{75.28} (1.95$\uparrow$) & \textbf{68.33} (3.53$\uparrow$) \\

\midrule
SATr\cite{li2022satr}&Baseline&25\% &-& Random &-&59.99    &68.05   &74.67   & 67.57\\

SATr+deweight\cite{jiang2014easy}&SCL\cite{jiang2014easy}&25\% &loss& Random & Hard Liner&58.17 (1.82$\downarrow$) &67.45  (0.60$\downarrow$) &73.84 (0.83$\downarrow$) & 66.49  (1.08$\downarrow$) \\

SATr+deweight\cite{liu2020self}&SCL\cite{liu2020self}&25\% &loss& Random &SPE&58.99 (1.00$\downarrow$) &67.87  (0.18$\downarrow$)&74.21 (0.46$\downarrow$)  & 67.02  (0.55$\downarrow$) \\

SATr+retrain\cite{Dong_2017_ICCV}&OHEM\cite{Dong_2017_ICCV}&25\% &loss& Random &$\times 2$  &61.71 (0.72$\uparrow$) &69.00  (0.95$\uparrow$)&75.37 (0.70$\uparrow$)  &  68.69 (1.12$\uparrow$) \\

SATr+$\hat{Mo}$+loss+uncer.&Ablation&25\% &loss+uncer.& hi+hi &-&65.17 (5.18$\uparrow$)&71.88 (3.83$\uparrow$) &77.30 (2.63$\uparrow$)  & 71.45 (3.88$\uparrow$)\\

SATr+Mo+l&Ablation&25\% &loss& hi+low &-&65.61 (5.62$\uparrow$)  &72.50 (4.45$\uparrow$)  &77.87  (3.20$\uparrow$) & 71.99 (4.22$\uparrow$) \\

SATr+Mo+u&Ablation&25\% &uncer.& hi+low &-&66.54 (6.65$\uparrow$) &73.87 (5.82$\uparrow$) &79.24 (4.57$\uparrow$) & 73.22 (5.65$\uparrow$)\\

SATr+Mo+l+u&Ours&25\% &loss+uncer.& hi+low &-&\textbf{68.54} (8.55$\uparrow$) &\textbf{75.38}  (7.33$\uparrow$) &\textbf{80.64} (5.97$\uparrow$)& \textbf{74.85} (7.28$\uparrow$) \\

\bottomrule[1pt]

A3D\cite{yang2021asymmetric}&Baseline &50\% &-& Random &-&72.52   &80.27 &86.14 &79.64 \\

A3D+deweight\cite{jiang2014easy}&SCL\cite{jiang2014easy} &50\% &loss& Random &Hard Liner&70.85 (1.67$\downarrow$)  &78.80 (1.47$\downarrow$)   &85.12 (1.02$\downarrow$)& 78.26 (1.38$\downarrow$) \\

A3D+deweight \cite{liu2020self}&SCL\cite{liu2020self} &50\% &loss& Random &SPE&72.31 (0.21$\downarrow$) &80.34 (0.07$\uparrow$) &86.01 (0.13$\downarrow$) & 79.55 (0.09$\downarrow$)\\

A3D+retrain\cite{Dong_2017_ICCV}&OHEM\cite{Dong_2017_ICCV} &50\% &loss& Random &$\times$2&73.07 (0.55$\downarrow$) &80.63 (0.36$\uparrow$) &86.24 (0.10$\uparrow$) & 79.98 (0.34$\uparrow$)\\

A3D+$\hat{Mo}$+l+u&Ablation&50\% &loss+uncer.& hi+hi &-&71.87 (0.65$\downarrow$) &79.45 (0.82$\downarrow$)  &85.60 (0.54$\downarrow$) & 78.97 (0.67$\downarrow$) \\

A3D+Mo+l+u&Ours&50\% &loss+uncer.& hi+low &-&\textbf{74.00} (1.48$\uparrow$)   &\textbf{81.23} (0.96$\uparrow$) &\textbf{86.48} (0.34$\uparrow$)  & \textbf{80.57} (0.93$\uparrow$) \\

\midrule
SATr \cite{li2022satr} &Baseline&50\% &-& Random &-&75.24 &82.19  &86.99 &81.47 \\

SATr+deweight\cite{jiang2014easy} &SCL\cite{jiang2014easy}&50\% &loss& Random & Hard Liner&74.63 (0.61$\downarrow$)&81.43 (0.76$\downarrow$)  &86.18 (0.81$\downarrow$) & 80.75 (0.72$\downarrow$)\\

SATr+deweight\cite{liu2020self} &SCL\cite{liu2020self}&50\% &loss& Random & SPE&75.19 (0.05$\downarrow$) &81.88 (0.31$\downarrow$)  &86.58 (0.41$\downarrow$) & 81.22 (0.25$\downarrow$)\\

SATr+retrain\cite{Dong_2017_ICCV} &OHEM\cite{Dong_2017_ICCV}&50\% &loss& Random & $\times$2&75.26 (0.02$\uparrow$) &82.17 (0.02$\downarrow$)  &86.41 (0.58$\downarrow$) & 81.28 (0.19$\downarrow$)\\

SATr+$\hat{Mo}$+l+u&Ablation&50\% &loss+uncer.& hi+hi &-&73.36 (1.88$\downarrow$) &80.52 (1.67$\downarrow$)  &85.40 (1.59$\downarrow$)& 79.76 (1.71$\downarrow$)\\

SATr+Mo+l&Ablation&50\% &loss& hi+low &-&74.52 (0.72$\downarrow$)  &81.83 (0.36$\downarrow$)  &86.69 (0.30$\downarrow$) & 81.01 (0.46$\downarrow$)\\

SATr+Mo+u&Ablation&50\% &uncer.& hi+low &-&75.69 (0.45$\uparrow$)&82.55 (0.36$\uparrow$)  &87.12 (0.13$\uparrow$) & 81.79 (0.32$\uparrow$) \\

SATr+Mo+l+u&Ours&50\% &loss+uncer.& hi+low &-&\textbf{76.97} (1.73$\uparrow$) &\textbf{83.66} (1.47$\uparrow$) &\textbf{87.27} (0.28$\uparrow$) & \textbf{82.63} (1.16$\uparrow$)\\
\toprule[1pt]
SATr\cite{li2022satr}&Baseline&100\% &-& Random &-&81.03 &86.64 &90.70 & 86.12 \\ 

SATr+deweight\cite{jiang2014easy}&SCL\cite{jiang2014easy}&100\% &loss& Random &Hard Liner&79.29 (1.74$\downarrow$) &85.38 (1.26$\downarrow$) &89.07 (1.63$\downarrow$) &  84.58 (1.54$\downarrow$)\\

SATr+deweight\cite{liu2020self}&SCL\cite{liu2020self}&100\% &loss& Random &SPE&80.40 (3.63$\downarrow$) &84.77 (1.87$\downarrow$)  &89.80 (0.9$\downarrow$)& 83.99 (2.13$\downarrow$) \\

SATr+retrain\cite{Dong_2017_ICCV}&OHEM\cite{Dong_2017_ICCV}&100\% &loss& Random &$\times$2&76.14 (4.89$\downarrow$) &83.12 (3.52$\downarrow$)  &88.03 (2.67$\downarrow$) & 82.43 (3.70$\downarrow$)\\

SATr+$\hat{Mo}$+l+u&Ablation&100\% &loss+uncer.& hi+hi &-&78.66 (2.37$\downarrow$) &85.18 (1.46$\downarrow$) &89.94 (0.76$\downarrow$) &84.59 (1.53$\downarrow$)\\

SATr+Mo+l&Ablation&100\% &loss& hi+low &-&80.10 (0.93$\downarrow$) &85.42 (1.22$\downarrow$)  &89.86 (0.84$\downarrow$) &85.13 (1.00$\downarrow$) \\

SATr+Mo+u&Ablation&100\% &uncer.& hi+low &-&80.91 (0.12$\downarrow$) &86.60 (0.04$\downarrow$) &90.53 (0.17$\downarrow$) &86.01 (0.11$\downarrow$)\\

SATr+Mo+l+u&Ours&100\% &loss+uncer.& hi+low &-&\textbf{81.96} (0.93$\uparrow$)&\textbf{87.97} (1.33$\uparrow$) &\textbf{91.36} (0.66$\uparrow$)  &\textbf{87.10} (0.97$\uparrow$)\\

\bottomrule[1pt]
A3D\cite{yang2021asymmetric} w/ GT ROI&Baseline &50\% &-& Random &-&93.45&95.63&97.22&98.39\\
SATr \cite{li2022satr} w/ GT ROI&Baseline &50\% &-& Random &-&94.04&96.00 &97.30&98.57\\
\bottomrule[1.5pt]

&&&&&\textbf{REVISED}&\textbf{TESTSET \cite{cai2020lesion}}&&&\\
\midrule

\textbf{Methods}&\textbf{Category}&\textbf{Data}&\textbf{Measurer}&\textbf{Sample}&\textbf{Deweight}&\textbf{$@0.5$}&\textbf{$@1$}&\textbf{$@2$}&Avg.[0.5,1,2]\\
\midrule
A3D\cite{yang2021asymmetric}&Baseline&25\% &-& Random &-& 77.34 & 82.50 &86.66&82.17\\

A3D+deweight\cite{jiang2014easy}&SCL\cite{jiang2014easy}&25\% &loss& Random &Hard Liner& 74.58 (2.76$\downarrow$)&80.29 (2.21$\downarrow$) & 85.22  (1.44$\downarrow$)&78.03 (4.14$\downarrow$)\\

A3D+deweight\cite{liu2020self}&SCL\cite{liu2020self}&25\% &loss& Random &SPE&75.75 (1.59$\downarrow$)& 81.54 (0.96$\downarrow$) & 86.02  (0.64$\downarrow$)& 81.10 (1.07$\downarrow$)\\

A3D+retrain\cite{Dong_2017_ICCV}&OHEM\cite{Dong_2017_ICCV}&25\% &loss& Random &$\times$2&77.47 (0.13$\uparrow$)& 82.38 (0.12$\downarrow$) & 86.76  (0.10$\uparrow$)&  82.20 (0.03$\uparrow$)\\

A3D+$\hat{Mo}$+l+u&Ablation&25\% &loss+uncer.& hi+hi &-&79.26 (1.92$\uparrow$)&84.60 (2.10$\uparrow$)& 87.66 (1.00$\uparrow$)&83.84 (1.67$\uparrow$)\\

A3D+Mo+l+u&Ours&25\% &loss+uncer.& hi+lo &-& \textbf{81.39} (4.05$\uparrow$)&\textbf{86.07} (3.57$\uparrow$) & \textbf{89.22} (2.56$\uparrow$)& \textbf{85.56} (3.39$\uparrow$)\\

\midrule
SATr\cite{li2022satr}&Baseline&25\% &-& Random &-& 75.87   & 79.92  &82.83  &  79.54\\

SATr+deweight\cite{jiang2014easy}&SCL\cite{jiang2014easy}&25\% &loss& Random & Hard Liner& 74.56 (1.31$\downarrow$) &78.82 (1.10$\downarrow$) & 81.88 (0.95$\downarrow$) & 78.42 (1.12$\downarrow$)\\

SATr+deweight\cite{liu2020self}&SCL\cite{liu2020self}&25\% &loss& Random &SPE &75.48 (0.39$\downarrow$) & 79.21 (0.71$\downarrow$) & 82.00 (0.83$\downarrow$) & 78.90 (0.64$\downarrow$)\\

SATr+retrain\cite{Dong_2017_ICCV}&OHEM\cite{Dong_2017_ICCV}&25\% &loss& Random &$\times$2 &75.37 (0.50$\downarrow$) & 79.11 (0.81$\downarrow$) & 82.26 (0.57$\downarrow$) &78.91 (0.63$\downarrow$)\\

SATr+$\hat{Mo}$+l+u&Ablation&25\% &loss+uncer.& hi+hi &-&79.33 (3.46$\uparrow$) & 83.80 (3.88$\uparrow$) &85.90 (3.07$\uparrow$) &  83.01 (3.47$\uparrow$)\\

SATr+Mo+l&Ablation&25\% &loss& hi+low &-& 78.53 (6.43$\uparrow$) & 83.47 (5.90$\uparrow$)  & 86.07 (5.45$\uparrow$) & 82.69 (3.15$\uparrow$)\\

SATr+Mo+u&Ablation&25\% &uncer.& hi+low &-&82.30 (4.67$\uparrow$)& 85.82  (4.67$\uparrow$)&88.28 (4.67$\uparrow$)& 85.47 (5.93$\uparrow$)\\

SATr+Mo+l+u&Ours&25\% &loss+uncer.& hi+low &-& \textbf{83.93} (8.06$\uparrow$)& \textbf{86.56} (7.64$\uparrow$) & \textbf{90.17} (7.34$\uparrow$)& \textbf{86.89} (7.35$\uparrow$)\\

\bottomrule[1pt]

A3D\cite{yang2021asymmetric} &Baseline&50\% &-& Random &-&86.09  & 88.93 & 91.21&88.74\\

A3D+deweight\cite{jiang2014easy}&SCL\cite{jiang2014easy} &50\% &loss& Random &Hard Liner& 85.08 (1.01$\downarrow$) & 88.09 (0.84$\downarrow$) & 90.41 (0.80$\downarrow$)& 87.86 (0.88$\downarrow$)\\

A3D+deweight \cite{liu2020self} &SCL\cite{liu2020self}&50\% &loss& Random &SPE&85.91 (0.18$\downarrow$) & 88.99 (0.06$\uparrow$) & 91.32 (0.11$\uparrow$)&  88.74 (-)\\

A3D+retrain\cite{Dong_2017_ICCV}  &OHEM\cite{Dong_2017_ICCV}&50\% &loss& Random &$\times$2&86.14 (0.05$\uparrow$) & 89.07 (0.14$\uparrow$) & 91.29 (0.08$\uparrow$)&   88.83 (0.09$\uparrow$)\\

A3D+$\hat{Mo}$+l+u&Ablation&50\% &loss+uncer.& hi+hi &-&85.81 (0.28$\downarrow$) & 88.39 (0.54$\downarrow$) & 90.41 (0.80$\downarrow$)& 88.20 (0.54$\downarrow$)\\

A3D+Mo+l+u&Ours&50\% &loss+uncer.& hi+low &-& \textbf{87.47} (1.40$\uparrow$)  & \textbf{90.27} (1.34$\uparrow$) & \textbf{91.80}  (0.59$\uparrow$) & \textbf{89.85} (1.11$\uparrow$)\\

\midrule
SATr \cite{li2022satr} &Baseline&50\% &-& Random &-& 86.94 & 90.35 & 92.96& 90.08\\

SATr+deweight\cite{jiang2014easy} &SCL\cite{jiang2014easy}&50\% &loss& Random & Hard Liner& 85.41 (1.53$\downarrow$)& 89.05 (1.30$\downarrow$) &92.13 (0.83$\downarrow$)&88.86 (1.22$\downarrow$)\\

SATr+deweight\cite{liu2020self}&SCL\cite{liu2020self} &50\% &loss& Random & SPE&86.02 (0.92$\downarrow$)&89.59 (0.76$\downarrow$) &92.84 (0.12$\downarrow$)& 89.48 (0.60$\downarrow$)\\

SATr+retrain\cite{Dong_2017_ICCV} &OHEM\cite{Dong_2017_ICCV} &50\% &loss& Random &$\times$2 &86.14 (0.80$\downarrow$)&89.66 (0.69$\downarrow$) &92.70 (0.26$\downarrow$)&89.50 (0.58$\downarrow$)\\

SATr+$\hat{Mo}$+l+u&Ablation&50\% &loss+uncer.& hi+hi &-&86.60 (0.34$\downarrow$)&90.07 (0.28$\downarrow$) &92.80 (0.16$\downarrow$)& 89.82 (0.26$\downarrow$)\\

SATr+Mo+l&Ablation&50\% &loss& hi+low &-&87.60 (0.66$\uparrow$) &90.87 (0.52$\uparrow$) & 93.40 (0.44$\uparrow$) &90.62 (0.54$\uparrow$)\\

SATr+Mo+u&Ablation&50\% &uncer.& hi+low &-&87.75 (0.81$\uparrow$)&91.11 (0.80$\uparrow$) &93.40 (0.44$\uparrow$)& 90.75 (0.67$\uparrow$)\\

SATr+Mo+l+u&Ours&50\% &loss+uncer.& hi+low &-& \textbf{88.41} (1.47$\uparrow$)& \textbf{91.50} (1.15$\uparrow$) & \textbf{94.03} (1.07$\uparrow$)& \textbf{91.31} (1.23$\uparrow$)\\

\bottomrule[1.5pt]
\end{tabular}

\end{lrbox}

\scalebox{0.77}{\usebox{\tablebox}}
\end{table*}

\section{Experiment}
\subsection{Dataset and Setting}
Our experiments are conducted on diverse datasets: DeepLesion\cite{yan18deeplesion} for ULD, Seg-C19 \cite{qiao2022semi}  for COVID CT segmentation, and CIFAR100-LT (imbalance rate = 0.01) \cite{zhou2020bbn} and CIFAR100-NL (human noise \cite{wei2021learning} and symmetric noise \cite{ma2020normalized})  for LT and NL image classification, respectively.

\textbf{The DeepLesion dataset} contains 32,735 lesions with a large diameter range (from 0.21 to 342.5mm) on 32,120 axial slices from 10,594 CT studies of 4,427 unique patients. The 12-bit intensity CT is rescaled to [0,255] with different window range settings and resized and interpolated according to the detection frameworks' settings. We follow the official split, i.e., $70\%$ for training, $15\%$ for validation, and $15\%$ for testing, with the testing set containing the official and revised \cite{cai2020lesion} version. To further testing the performance on a small dataset, we also conduct experiments using 25\% and 50\% training data. The number of false positives per image (FPPI) is used as the evaluation metric. For training, we use the original network settings. As for the loss selection, we use the anchor classification loss in Region Proposal Network (RPN) for data difficulty measurement. As for the uncertainty calculation, the disturbances ($G=8$) are added to the feature maps after the first CNN block of the detector backbone, and the uncertainty of RPN classification feature maps is taken as the uncertainty. Each pixel value of $t^{g}$ is sampled from a uniform distribution with $\gamma = 0.3$.

\textbf{The Seg-C19 dataset} is a COVID-19 lesion segmentation dataset containing  908 annotated CT slices from 35 patients \cite{qiao2022semi}. We use 724, 184 and
355 slices for training, validation, and testing. The training, validation, and test sets come from different patients.
Three different windows (i.e., [-174, 274], [-1493, 484], and [-534, 1425]) are used to convert
12-bit CT images into three-channel images and normalize the values of each channel, respectively. All the images in both training and testing sets are resized to 512 × 512. To evaluate the robustness of our method under different training set sizes, we use three different training set sizes, i.e., 72 (10\%), 352 (50\%), and 724 (100\%)  images.

The CIFAR-100 dataset \cite{krizhevsky2009learning}, a subset of the Tiny Images dataset, consists of 60,000 32 $\times$ 32 color images. There are 500 training images and 100 test images per class. The CIFAR100-LT (imbalance rate = 0.01) \cite{zhou2020bbn} dataset, CIFAR100-NL (human noise, noise rate = 0.42) \cite{wei2021learning} dataset, and CIFAR100-NL (symmetric noise, noise rate = 0.4) \cite{ma2020normalized} are both build based on CIFAR-100.

\textbf{The CIFAR-100 LT} dataset is a long-tailed version of CIFAR-100, specifically designed to study and address the challenges posed by class imbalance in machine learning and computer vision tasks. The dataset is created by reducing the training samples per class according to an exponential function.

\textbf{The CIFAR-100 NL-human noise} dataset is a dataset for noisy label learning, which consists of artificially introduced noises in the training data labels. Here we use an official version by \cite{wei2021learning}, which consists of 42\% noise labels.

\textbf{The CIFAR-100 NL-symmetric noise} is another dataset for noisy label learning with symmetric noises. we follow \cite{ma2020normalized} and \cite{wei2021learning}, using human noise and symmetric noise labels in the training set.


It is worth noting that the testing set for the latter three tasks remains the same as the original CIFAR-100.

\subsection{Loss and Uncertainty}

For the two-stage ULD methods, there are at least four different losses: the RPN anchor classification loss and RPN box regression loss in Stage 1, and Region of Interest (RoI) box classification loss and regression loss in Stage 2. We need to select appropriate losses and feature maps (for uncertainty estimation) to measure data difficulty. In our work, we adopt a `fixed one, test another' strategy to evaluate the performance of two key components in two stages, i.e., RPN in Stage 1 and ROI classification and regression in Stage 2. We first train the network with its original architectures and experimental settings to obtain a well-trained model weight, and then we replace the RoIs with GT Bounding Boxes (BBoxes) during the test stage.  We report the experimental result in Table \ref{results} for two SOTA two-stage ULD methods based on 50\% training data. When the RoIs are replaced with the GT BBoxes, a significant performance improvement is observed compared to the original approach, indicating that Stage 1 is more appropriate for measuring data difficulty. Hence, we use the RPN anchor classification loss as the difficulty measure loss and use the RPN classification feature maps for uncertainty calculation.

For the COVID-19 lesion segmentation, LT classification, and NL classification tasks, we directly use their loss as a difficulty measurer and introduce disturbances into the feature map of the bottleneck to obtain the uncertainty estimation.

\begin{table*}[!h]
 \centering
 \caption{Top-1 accuracy of 7 baselines on CIFAR-100-LT \cite{zhou2020bbn} with an imbalance ratio of 0.01 with different batch sizes (BS).} \label{CIFAR100_LT}
 \begin{lrbox}{\tablebox}
 \begin{tabular}{lllllll}
   \toprule[1.5pt]
   Method&Measurer&Uncertainty manner&Sampling& \multicolumn{3}{c}{Accuracy}\\
   &&&&       BS=64&BS=32&BS=16\\
   \toprule[1pt]
   R32\cite{targ2016resnet}&-&-&random&0.284& 0.311& 0.314\\
   R32+SCL\cite{jiang2014easy}&loss&-&random&0.291(0.7\%$\uparrow$)& 0.297(-1.4\%$\downarrow$)& 0.300(-1.4\%$\downarrow$)\\
   R32+SCL\cite{liu2020self}&loss&-&random&0.289(0.5\%$\uparrow$)& 0.307(-0.4\%$\downarrow$)& 0.311(-0.3\%$\downarrow$)\\
   R32+OHEM\cite{Dong_2017_ICCV}&loss&-&random&0.287(0.3\%$\uparrow$)& 0.301(-1.0\%$\downarrow$)& 0.319(0.5\%$\uparrow$)\\
   R32+Mo+l&loss&-&low+hi&0.342(5.8\%$\uparrow$)& 0.326(1.5\%$\uparrow$)& 0.315(0.1\%$\uparrow$)\\
   R32+Mo+u(w/o disturbance)&uncer.&1 disturbance&low+hi&0.318(3.4\%$\uparrow$)& 0.336(2.5\%$\uparrow$)& 0.326(1.2\%$\uparrow$)\\
   R32+Mo+u(w/ disturbance)&uncer.&8 disturbances&low+hi&0.300(1.6\%$\uparrow$)& 0.314(0.3\%$\uparrow$)& 0.332(1.8\%$\uparrow$)\\
   R32+Mo+u+l(ours w/o disturbance)&loss+CAM.&-&low+hi&0.332(4.8\%$\uparrow$)& 0.341(3.0\%$\uparrow$)& 0.357(4.3\%$\uparrow$)\\
   R32+Mo+u+l(ours w/o disturbance)&loss+uncer.&1 disturbance&low+hi&0.351(6.7\%$\uparrow$)& 0.353(4.2\%$\uparrow$)& 0.363(4.9\%$\uparrow$)\\
   R32+Mo+u+l(ours w/ disturbance)&loss+uncer.&8 disturbances&low+hi&0.313(2.9\%$\uparrow$)& 0.358(4.7\%$\uparrow$)& 0.367(5.3\%$\uparrow$)\\

   \cline{1-7}
 Focal loss($\gamma=2$)\cite{lin2017focal}&-&-&random&0.314& 0.341& 0.356\\
 Focal loss($\gamma=2$)+ours&loss+uncer.&8 disturbances&low+hi&0.348(3.4\%$\uparrow$)& 0.343(0.2\%$\uparrow$)& 0.376(2.0\%$\uparrow$)\\
   \cline{1-7}
 Imbal. loss\cite{cui2019class} &-&-&random&0.300& 0.326& 0.321\\
 Imbal. loss+ours &loss+uncer.&8 disturbances&low+hi&0.341(4.1\%$\uparrow$)& 0.328(0.2\%$\uparrow$)& 0.337(1.6\%$\uparrow$)\\
    \cline{1-7}
 GGD\cite{han2021general}&-&&random&0.318& 0.310& 0.327\\
 GGD+ours&loss+uncer.&8 disturbances&low+hi&0.347(2.9\%$\uparrow$)& 0.351(4.1\%$\uparrow$)& 0.368(4.1\%$\uparrow$)\\
    \cline{1-7}
 IB32\cite{park2021influence}&-&&random&0.425& 0.422& 0.421\\
 IB32+ours&loss+uncer.&8 disturbances&low+hi&0.439(1.4\%$\uparrow$)& 0.431(0.9\%$\uparrow$)& 0.430(0.9\%$\uparrow$)\\
    \cline{1-7}
 LDAM\cite{cao2019learning}&-&&random&0.409& 0.402& 0.387\\
 LDAM+ours&loss+uncer.&8 disturbances&low+hi&0.410(0.1\%$\uparrow$)& 0.411(0.9\%$\uparrow$)& 0.433(4.6\%$\uparrow$)\\
    \cline{1-7}
 GCL-stage1\cite{li2022long}&-&&random&0.458& 0.466& 0.459\\
 GCL-stage1+ours&loss+uncer.&8 disturbances&low+hi&\textbf{0.468}(1.0\%$\uparrow$)& \textbf{0.475}(0.9\%$\uparrow$)& \textbf{0.469}(1.0\%$\uparrow$) \\

   \bottomrule[1.5pt]
 \end{tabular}
 \end{lrbox}
 \scalebox{1}{\usebox{\tablebox}}
\end{table*}

\begin{table*}[!h]
\centering
\caption{UPPER: Top-1 accuracy of ResNet-32 (R32) on CIFAR-100-NL with human noise of 0.42 noise rate \cite{wei2021learning} under different batch sizes (BS). LOWER: Top-1 accuracy of 5 baselines on CIFAR-100-LT with symmetric noise of 0.4 noise rate \cite{ma2020normalized} under different batch sizes (BS).} \label{CIFAR100_NL}
\begin{lrbox}{\tablebox}
\begin{tabular}{lllllll}
  \toprule[1.5pt]
Method & Noise type & Measurer & Uncertainty manner & Sampling & \multicolumn{2}{c}{Accuracy} \\
&&&&   &    BS=32 & BS=16 \\
\toprule[1pt]
R32\cite{targ2016resnet} & \multirow{6}{*}{Human noise} & - & - & random & 0.534 & 0.504 \\
R32+deweight\cite{jiang2014easy} &&loss&-&random& 0.537 (0.3\%$\uparrow$)&0.525 (2.1\%$\uparrow$)\\
R32+deweight\cite{liu2020self}&&loss&-&random& 0.521 (1.6\%$\downarrow$)&0.524 (2.0\%$\uparrow$)\\
R32+Mo+l&&loss&-&low+hi&0.559 (2.5\%$\uparrow$) & 0.533 (2.9\%$\uparrow$) \\
R32+Mo+u(w/ disturbance)&&uncer.&8 disturbances&low+hi& 0.560 (2.6\%$\uparrow$) & 0.541 (3.7\%$\uparrow$) \\
R32+Mo+u+l(ours w/ disturbance)&&loss+uncer.&8 disturbances&low+hi& 0.564 (3.0\%$\uparrow$) & 0.550 (4.6\%$\uparrow$) \\
\toprule[1pt]
Focal loss($\gamma=0.5$)\cite{lin2017focal}&\multirow{7}{*}{Symmetric noise}&-&-&random&0.487 &0.507 \\
Focal loss($\gamma=0.5$)+ours&&-&-&random&0.505 (1.8\%$\uparrow$) &0.521  (1.4\%$\uparrow$)\\
\cline{1-1}\cline{3-7}
NLNL\cite{kim2019nlnl}&&-&-&random&0.414 & 0.427 \\
SCE\cite{wang2019symmetric}&&-&-&random&0.432 &0.443 \\
\cline{1-1}\cline{3-7}
GCE\cite{zhang2018generalized}&&-&-&random&0.590 & 0.610 \\
GCE+ours&&loss+uncer.&8 disturbances&low+hi&\textbf{0.594} (0.4\%$\uparrow$) &\textbf{0.621} (1.1\%$\uparrow$) \\
\cline{1-1}\cline{3-7}
NECE+RCE\cite{ma2020normalized}&&-&-&random&0.573 &0.584 \\
NECE+RCE+ours&&loss+uncer.&8 disturbances&low+hi&0.581 (0.8\%$\uparrow$)&0.602 (1.8\%$\uparrow$)\\
\bottomrule[1.5pt]
\end{tabular}
\end{lrbox}
\scalebox{1}{\usebox{\tablebox}}
\end{table*}

\begin{table}[!h]
  \centering
\caption{2D CT segmentation performance with various amounts of training samples from COVID dataset Seg-C19\cite{qiao2022semi}.}
\begin{lrbox}{\tablebox}
\begin{tabular}{p{22mm}p{18mm}p{18mm}p{18mm}}

  \toprule[1.5pt]
\multirow{3}{*}{Method}
&\multicolumn{3}{c}{Dice. (p value)} \\ \cline{2-4}
 &\multicolumn{3}{c}{Number of training CT slices}\\ 
 &72&352&724\\ \hline

DenseUNet~\cite{li2018h} &.6640&.6733&.6890\\
COPLE-Net~\cite{wang2020noise} &.6465&.7067&.7094\\
Inf-Net~\cite{fan2020inf}&.6683&.7162&.7244\\
U-Net~\cite{ronneberger2015u} &.6670&.6909&.7193\\
U-Net+ours &.6700 (0.017)&.7058 (0.028)&.7278 ($\textless$0.01)\\
nnUNet~\cite{isensee2021nnu} &.6689 &.7125&.7255\\
nnUNet+ours &\textbf{.6728} ($\textless$ 0.01) &\textbf{.7177} (0.033) &\textbf{.7357} ($\textless$0.01)\\
  \toprule[1.5pt]

\end{tabular} \label{table:C19_performance}
\end{lrbox}
\scalebox{0.95}{\usebox{\tablebox}}
\end{table}

\subsection{ULD on DeepLesion}
Two SOTA ULD approaches \cite{li2021conditional,li2022satr} are compared to evaluate MoMBS's effectiveness via the original testing set and revised testing set from \cite{cai2020lesion}. All results in Table \ref{results} are obtained with batchsize = 4 because the SOTA baseline results are also archived under these settings. The influence of batchsize is discussed in \ref{ablaion}.

\textbf{Partial training results.} As shown in Table \ref{results}, under the $25\%$ and $50\%$ training data settings, the deweighting methods, i.e., SCL and OHEM, are harmful to network training. The anti-Mo methods, which pairs low- (or high-) difficulty with low- (or high-) difficulty data, can bring performance improvement as the mechanism of pairing \{$<u^l,l^h>$,$<u^h,l^l>$\} together influences each other, but causes a less effect on $<u^h,l^h>$ or $<u^l, l^l>$. This advantage also shrinks with more training data is used.
The loss-based MoMBS methods bring improvement in the  $25\%$ training data setting but fail in the $50\%$  training data setting, while the uncertainty-based MoMBS methods still produce marginal performance improvement in the $50\%$ training data setting. When combining them to form MoMBS produces the optimal result, which also shows the drawbacks of the methods that use a single difficulty measurer.

\textbf{Full training results.}  As shown in Table \ref{results}, the proposed MoMBS follows a similar rule in partial training, but the improvements under full training setting become marginal along with an  increased training set size.

\subsection{COVID Lesion Segmentation on Seg-C19}
We report the COVID lesion segmentation results on the Seg-C19 dataset \cite{qiao2022semi} in Table \ref{table:C19_performance}. We demonstrate the effectiveness of MoMBS compared to two SOTA segmentation methods, using 72, 352, and 724 CT training slices, respectively. In alignment with the official network settings of the six SOTA methods, we use a batch size of 4. Given that the test set is comparatively smaller than that of the other four tasks, we also include p-value results. As indicated in Table \ref{table:C19_performance}, all p-values are below 0.05, indicating that our method can significantly improve the baseline models.

\subsection{LT Image Classification on CIFAR100-LT}
We report the results on CIFAR100-LT (imbalance ratio=0.01) \cite{zhou2020bbn}, a well-known LT benchmark classification dataset, to demonstrate the generality of MoMBS. As shown in Table \ref{CIFAR100_LT}, all ResNet-32 results are improved with our Mo sampling. Especially, with our MoMBS, the top-1 ACC improvement of \textcolor{blue}{0.67/0.47/0.53} are realized under the 64/32/16 batch size, respectively. Besides, 6 SOTA LT methods are also improved. It is worth noting that this task requires less GPU memory per image and the best results are obtained with relatively large batchsizes.  Hence, our results are demonstrated on a relatively larger batchsize.

\subsection{NL Image Classification  on CIFAR100-NL}
We present the results on CIFAR100-NL (human noise, noise rate=0.42) and CIFAR100-NL (symmetric noise, noise rate=0.4), two recognized benchmarks for NL classification datasets, to further illustrate the versatility of MoMBS. As depicted in Table. \ref{CIFAR100_NL}, all ResNet-32 results are improved by employing our Mo sampling and MoMBS. Improvements in top-1 ACC of \textcolor{blue}{0.3/0.46} are achieved under the 32/16 batch size, respectively.  Additionally, our approach also advances 3 SOTA NL approaches. It should be emphasized that the inconsistency in the NL dataset is due to our adherence to the official settings of various methods, which is crucial for achieving the reported performance.

\subsection{Ablation Study}\label{ablaion}

We provide ablation study for the key components in our approach, i.e., sample difficulty assessor, and sampling method.  We also evaluate the effect of varying the numbers of batch sizes and pivot epochs to the performance.

\textbf{Sample difficulty assessor:} As indicated in Tables \ref{results}, \ref{CIFAR100_LT}, and \ref{CIFAR100_NL}, incorporating uncertainty into the sample difficulty assessment enhances performance across all three tasks.


\textbf{Sampling method:} As demonstrated in Tables \ref{results}, \ref{CIFAR100_LT}, and \ref{CIFAR100_NL}, maintaining an even total minibatch difficulty across all minibatches proves superior to other sampling methods, such as random sampling or pairing low-difficulty samples together.

\begin{figure}[!h]
 \centering
 \caption{Ablation study for Batchsize (BS) and pivot epoch based on DeepLesion \cite{cai2020lesion}, CIFAR100-LT \cite{zhou2020bbn}, CIFAR100N with human noise \cite{wei2021learning} and CIFAR100N with symmetric noise \cite{ma2020normalized}. } \label{batchsize_results}
 \begin{lrbox}{\tablebox}
 \begin{tabular}{p{12mm}p{28mm}<{\centering}p{25mm}<{\centering}p{8mm}<{\centering}p{8mm}<{\centering}p{8mm}<{\centering}}

 \bottomrule[1.5pt]
     \textbf{Methods}&\textbf{Data}&\textbf{Metrics}&\textbf{BS=8}&\textbf{BS=4}&\textbf{BS=2}\\
   \toprule[1pt]

   SATr&25\% DeepLesion &Avg. FP[0.5,1,2]& 67.99 &67.57&66.49 \\
   SATr+ours&25\% DeepLesion &Avg. FP[0.5,1,2]&\textbf{72.22}&\textbf{74.85}&\textbf{74.79} \\
   \bottomrule[1.5pt]
 \end{tabular}
 \end{lrbox}
 \scalebox{0.75}{\usebox{\tablebox}}

 \par\vspace{1ex}

 \begin{lrbox}{\tablebox}
 \begin{tabular}{p{15mm}p{28mm}<{\centering}p{20mm}<{\centering}p{8mm}<{\centering}p{8mm}<{\centering}p{8mm}<{\centering}}

 \bottomrule[1.5pt]
     \textbf{Methods}&\textbf{Data}&\textbf{Metrics}&\textbf{BS=64}&\textbf{BS=32}&\textbf{BS=16}\\
   \toprule[1pt]

   R32&CIFAR100-NL& Top-1 Acc.& 54.1 &53.4&50.4 \\
    R32+ours&CIFAR100-NL& Top-1 Acc.& \textbf{55.1} &\textbf{56.4}&\textbf{55.0} \\
   \toprule[1pt]
   R32&CIFAR100-LT& Top-1 Acc.& 28.4 &31.1&31.4 \\
    R32+ours&CIFAR100-LT& Top-1 Acc.& \textbf{31.3} &\textbf{35.8}&\textbf{36.7} \\

   \bottomrule[1.5pt]
 \end{tabular}
 \end{lrbox}
 \scalebox{0.75}{\usebox{\tablebox}}

 \par\vspace{1ex}

 \begin{lrbox}{\tablebox}
 \begin{tabular}{p{10mm}p{23mm}<{\centering}p{8mm}<{\centering}p{8mm}<{\centering}p{8mm}<{\centering}p{8mm}<{\centering}p{8mm}<{\centering}p{8mm}<{\centering}}

 \bottomrule[1.5pt]
     \textbf{Methods}&\textbf{Data}&\textbf{PE=$\infty$}&\textbf{PE=5}&\textbf{PE=10}&\textbf{PE=20}&\textbf{PE=30}&\textbf{PE=50}\\
   \toprule[1pt]
   R32+ours&CIFAR100-NL&50.4 & \textbf{55.0} &54.6&53.8&54.9&51.0 \\
      \toprule[1pt]
    R32+ours&CIFAR100-LT&31.4 &32.9 &36.3&\textbf{36.7}&33.4&31.8 \\

   \bottomrule[1.5pt]
 \end{tabular}
 \end{lrbox}
 \scalebox{0.75}{\usebox{\tablebox}}

 \par\vspace{1ex}

 \begin{lrbox}{\tablebox}
 \begin{tabular}{p{15mm}p{25mm}<{\centering}p{7mm}<{\centering}p{7mm}<{\centering}p{7mm}<{\centering}p{7mm}<{\centering}p{7mm}<{\centering}p{7mm}<{\centering}}

 \bottomrule[1.5pt]
     \textbf{Methods}&\textbf{Data}&\textbf{PE=$\infty$}&\textbf{PE=30}&\textbf{PE=40}&\textbf{PE=50}&\textbf{PE=60}&\textbf{PE=70}\\
   \toprule[1pt]
   SATr+ours&25\% DeepLesion&67.57&70.11&71.21&\textbf{74.85}&72.54&71.99\\
   \bottomrule[1.5pt]
 \end{tabular}
 \end{lrbox}
 \scalebox{0.75}{\usebox{\tablebox}}

 \includegraphics[scale=0.53]{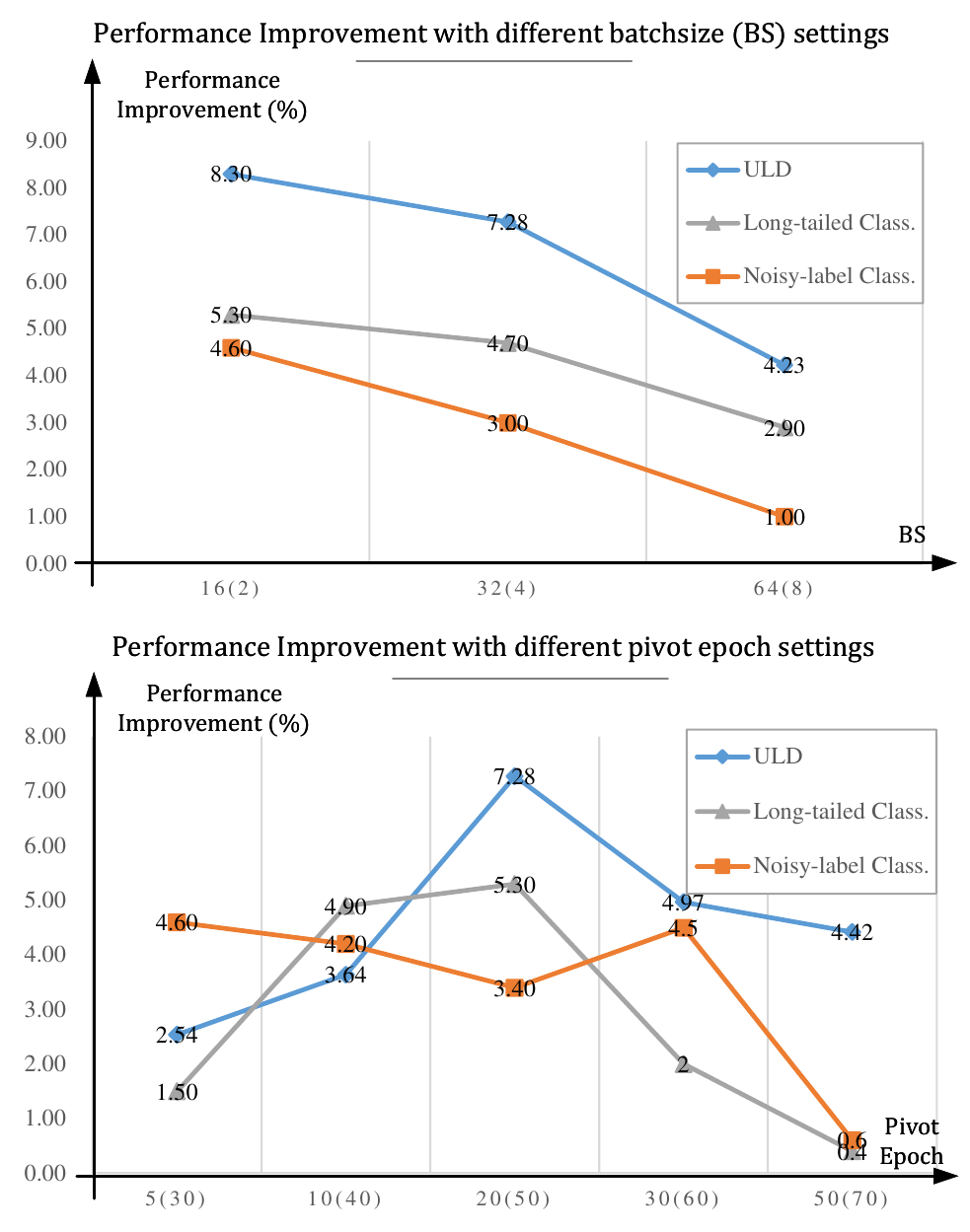}
 \caption{Performance improvements with different batchsizes settings (UPPER) and pivot epoch settings (LOWER). }\label{batchsize_results_fig}

\end{figure}

\textbf{Batch size:} As evidenced in Fig. \ref{batchsize_results} and Fig. \ref{batchsize_results_fig}, generally, a larger batch size tends to slightly diminish the effectiveness of MoMBS. Smaller batch sizes, in contrast,  prove more suitable for MoMBS. Due to constraints related to GPU memory, the results for batch sizes greater than 8 for ULD tasks cannot be provided.

\textbf{Pivot epoch:} As illustrated in Table \ref{batchsize_results} and Fig. \ref{batchsize_results_fig}, setting the pivot epoch too large or too small compromises the effectiveness of MoMBS. Employing MoMBS too late increases the challenges for the method to escape the local minimum, whereas activating MoMBS too early introduces instability issues of loss and uncertainty.

\begin{figure*}[!h]
\centering
\setlength{\abovecaptionskip}{0.cm}
\includegraphics[scale=0.6]{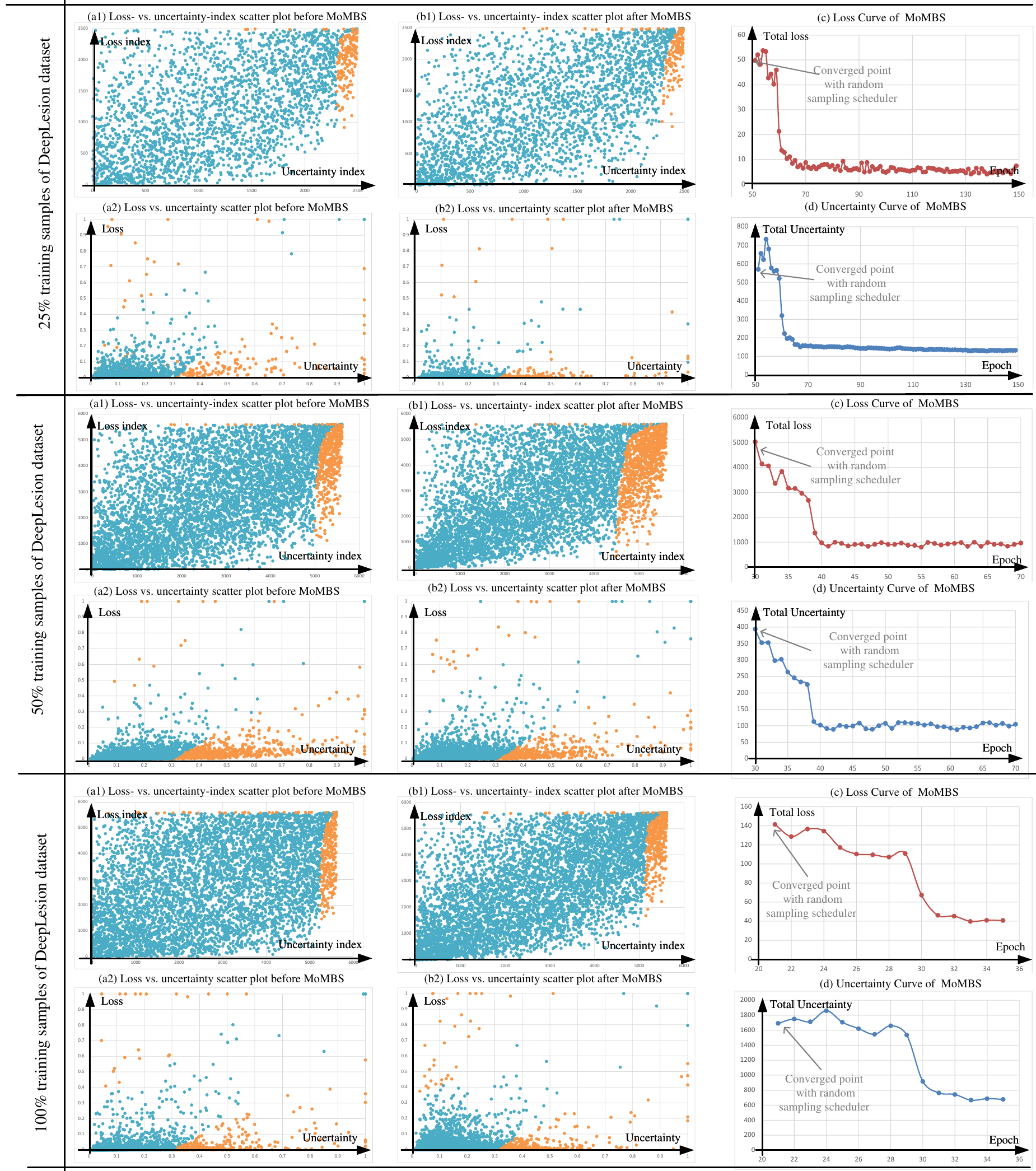}
\caption{Illustration of loss and uncertainty relationship based on \cite{li2022satr}. Yellow (or cyan) denotes the sample whose absolute difference between uncertainty and loss is greater (or less) than 0.3.}\label{Fig3_loss_uncertainty}

\end{figure*}

 \begin{figure}[!h]
 \centering
 \setlength{\abovecaptionskip}{0.cm}
 \includegraphics[scale=0.48]{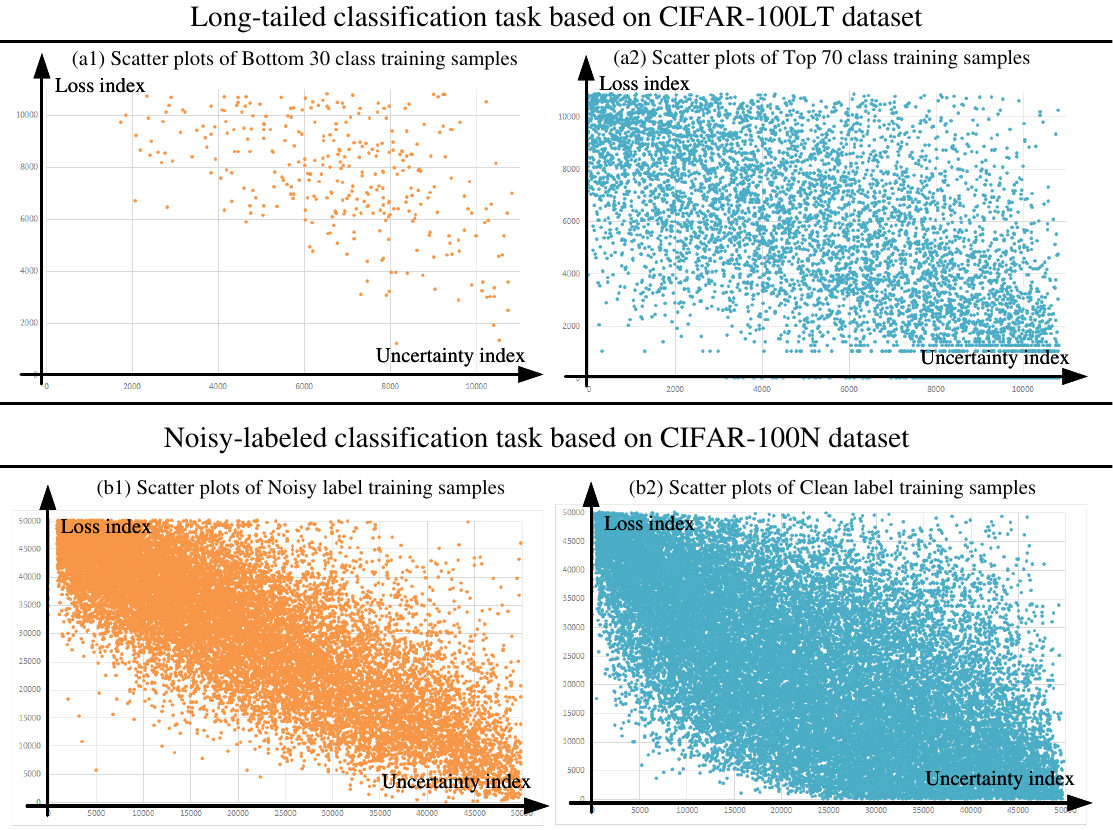}
 \caption{UPPER: Loss vs. uncertainty scatter plots of LT samples (a1) and other samples (a2). LOWER: Loss vs. uncertainty scatter plots of NL samples (b1) and clean samples (b2). }\label{Fig4_loss_uncertainty_more}

\end{figure}

\begin{figure}[!h]
\centering
\setlength{\abovecaptionskip}{0.cm}
\includegraphics[scale=0.3]{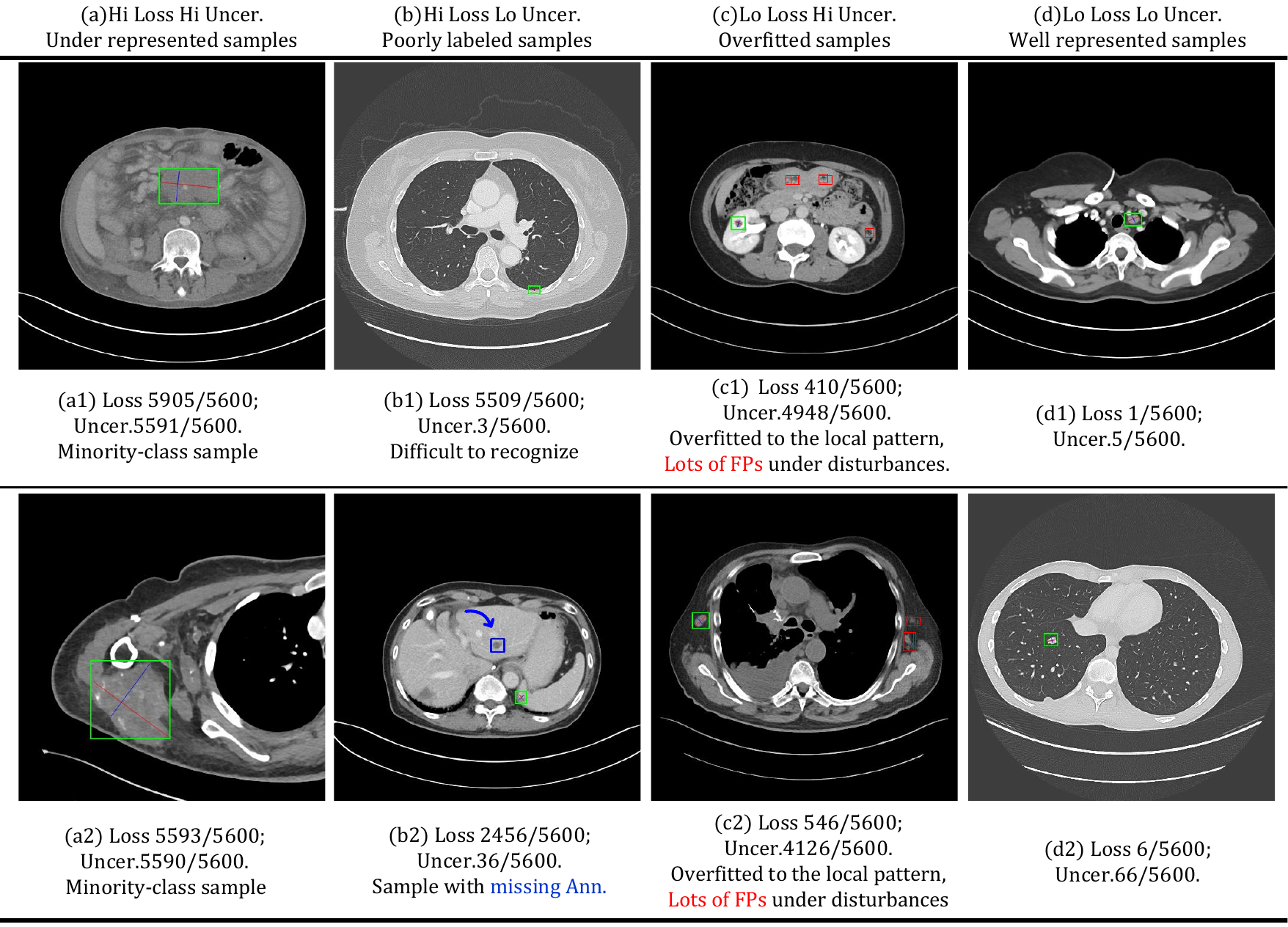}
\caption{Eight samples from four data types.   Based on loss-based difficulty measures, samples (a) and samples (b) are grouped as hard samples, while (c) and (d) are grouped as easy samples. Our MoMBS further sub-categorizes hard samples into (a) under represented images and  (b) poorly labeled samples, and distinguishes (c) overfitted samples from easy samples. Through minibatch pairing according to these categories, MoMBS effectively manages these samples. The ineffective situation of MoMBS is distinguishing
b1 as the poorly labeled sample while it is an under represented sample.   MoMBS has a minor effect on b1's training and we believe that improved network design should be a future direction to handle this.}
\label{suppFig2_samples}
\end{figure}

\subsection{Visualization}
In this part, we give out visual results to substantiate the superiority of MoMBS. We introduce visual results for ULD, LT classification, and NL classification tasks respectively.

\subsubsection*{\bf Visualization for ULD}

We here provide visual results for ULD in 3 respects: 1)  Visualization of samples from DeepLesion with different uncertainty and loss, 2) Illustration of loss and uncertainty relationship based on a loss vs. uncertainty scatter plot, and 3) Loss- and uncertainty- maps along with training epochs.

 \underline{Visualization of samples.}
In Fig. \ref{suppFig2_samples}, we present eight samples to further demonstrate the efficacy of our difficulty measure mechanism. For (a), the minority-class samples a1 and a2 are accurately identified as under represented, requiring a greater attention from the network. In (b), while b2 is correctly identified as a poorly labeled sample, b1 is mistakenly grouped with b1. For such instances, refining the network design might offer a better approach instead of our proposed MoMBS. In (c), both two samples are overfitted samples, and a little disturbance largely influences their prediction. Given their low loss values, additional loss gradient descent training on them offers a limited improvement. Lastly, in (d), they are well represented samples.

 \underline{Relationship between loss and uncertainty}
In order to underscore the importance of integrating uncertainty in data difficulty estimation, we provide empirical evidence to support our argument. Specifically, we compute the loss and uncertainty for all training samples using two SOTA ULD methods with  25\% training data and plot the data in 2D dashed plots. We hereby show the results based on \cite{li2022satr} in Fig. \ref{Fig3_loss_uncertainty}. Our results reveal several key insights:
(1) Fig. \ref{Fig3_loss_uncertainty} (a)$\&$(b) illustrate the low correlation between loss and uncertainty. We observe that uncertainty values are more scattered compared to loss values, which tend to concentrate on small values. This divergence may be attributed to network training's direct influence on the loss gradient while leaving the uncertainty gradient less affected. (2) The index-based method can eliminate singular points in the value-based map,  highlighting that our approach effectively circumvents fluctuations in network training and issues of incomparability between the loss and uncertainty scales. (3) Fig. \ref{Fig3_loss_uncertainty} (a1)$\&$(b1) illustrate that more samples concentrate on the well represented and under represented areas after using MoMBS, which is consistent with the observation that the majority of training samples have correct labels.

Moreover, according to Fig. \ref{Fig3_loss_uncertainty} (c)$\&$(d), we observe that MoMBS can further reduce the total loss across the entire training dataset after the network converges (the epoch reaches the best performance without MoMBS), 
which suggests that maintaining minibatch difficulty is useful for the network to find an effective convergence direction. MoMBS also shows a strong capacity to reduce uncertainty for ULD. This indicates that the network trained with MoMBS is more robust against disturbance and more reliable.

\subsubsection*{\bf Visualization for LT and NL image classification}

From our earlier discussion, it is evident that CIFAR100-LT and CIFAR100-NL exhibit significant challenges due to their extreme long-tailed and noisy-label issues, respectively. As a result, they align less with our proposed sample categorization compared to datasets like DeepLesion and Seg-C19. This discrepancy occurs because the network must converge over a large number of long-tailed classes or noisy-label samples, which can substantially influence the training of other samples.

Given that CIFAR100-LT and CIFAR100-NL are manually derived from the original CIFAR-100 dataset, it is feasible to separate the long-tailed class and noisy-labeled training samples. This separation allows us to study MoMBS's effectiveness on these samples. For the LT task, we present the loss vs. uncertainty graph for the bottom 30 class samples and the other 70 class samples in panel (a) of Fig. \ref{Fig4_loss_uncertainty_more}. It is evident that while the long-tailed class training samples, which should be considered as under represented, do not strictly adhere to the categorization, they predominantly occupy the top-right section. This positioning suggests a trend of partial alignment with our proposed categorization.

Regarding the NL task, we illustrate the loss vs. uncertainty map for noisy-labeled samples and clean samples in panel (b) of Fig. \ref{Fig4_loss_uncertainty_more}. The noisy-labeled samples generally conform to our proposed categorization, with most being identified as poorly labeled. Meanwhile, the clean samples exhibit a trend of following the categorization more closely.

\section{Conclusions and Future Work}
This paper contends that effective minibatch sampling is crucial for tasks with diverse-quality training samples like ULD, and long-tailed and noisy-labeled image classification. To address this challenge,  we introduce a novel MBS strategy called MoMBS. It incorporates both loss and uncertainty rank scores to obtain a more accurate estimate of sample difficulty and then employs mixed-order sampling to mitigate sample under-utilization and unnecessary data conflict, thus bringing performance improvement.  We validate the efficacy of MoMBS through experimental explanation and comprehensive experiments on ULD, COVID19 CT segmentation, long-tailed image classification, and noisy-labeled image classification.
This efficacy is particularly pronounced in scenarios with a limited number of training samples and a reasonable proportion of low-quality samples.

In the future, we plan to explore an even better MBS strategy. Currently we use uncertainty as a part of solution, but there might be other better solutions. We have experimented with using Classification Activation Maps as a substitute for uncertainty, but this did not yield substantial enhancements as shown in Table \ref{CIFAR100_LT}. Further, it is not clear whether mixed-order sampling can be improved.


\ifCLASSOPTIONcaptionsoff
  \newpage
\fi

\bibliographystyle{unsrt}
\bibliography{egbib}

\begin{IEEEbiography}[{\includegraphics[width=1in,height=1.25in,clip,keepaspectratio]{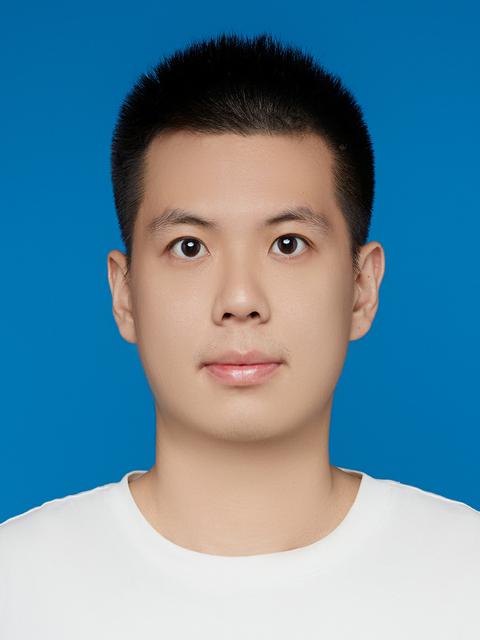}}]{Han Li}
received the B.E. degree from Henan Polytechnic University (HPU) in 2016, and the M.S. degree in computer science from the Institute of Computing Technology (ICT), University of Chinese Academy of Sciences (UCAS) in 2021. He is currently pursuing the Ph.D. degree from the School of Biomedical Engineering \& Suzhou Institute for Advanced Research, University of Science and Technology of China (USTC). His research interests include computer vision, machine learning and image processing, with applications to biomedical engineering.
\end{IEEEbiography}

\begin{IEEEbiography}[{\includegraphics[width=1in,height=1.25in,clip,keepaspectratio]{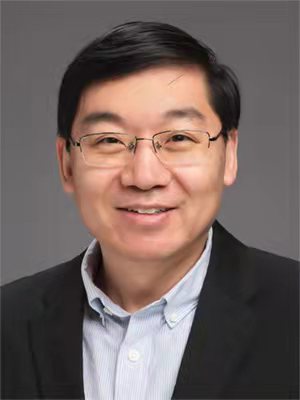}}]{Prof. Hu Han}
(Member, IEEE) received the Ph.D. degree in computer science from the Institute of Computing Technology (ICT), Chinese Academy of Sciences (CAS), Beijing, China, in 2011. He is a Professor with ICT, CAS. He was a Research Associate with PRIP Lab, Michigan State University, East Lansing, MI, USA, and a Visiting Researcher with Google, Mountain View, CA, USA, from 2011 to 2015. He has published more than 70 papers in journals and conferences, including IEEE TRANSACTIONS ON PATTERN ANALYSIS AND MACHINE INTELLIGENCE, IEEE TRANSACTIONS ON IMAGE PROCESSING, IEEE TRANSACTIONS ON INFORMATION FORENSICS AND SECURITY, IEEE TRANSACTIONS ON BIOMETRICS, BEHAVIOR, AND IDENTITY SCIENCE, Pattern Recognition, CVPR, NeurIPS, ECCV, and MICCAI, with more than 4000 Google Scholar citations. His research interests include computer vision, pattern recognition, and biometrics. Dr. Han was a recipient of the 2020 IEEE Signal Processing Society Best Paper Award, the 2019 IEEE FG Best Poster Presentation Award, and the 2016/2018 CCBR Best Student/Poster Award. He is/was an Associate Editor of Pattern Recognition, the Area Chair of ICPR2020, and a Senior Program Committee Member of IJCAI2021.
\end{IEEEbiography}

\begin{IEEEbiography}
[{\includegraphics[width=1in,height=1.25in,clip,keepaspectratio]{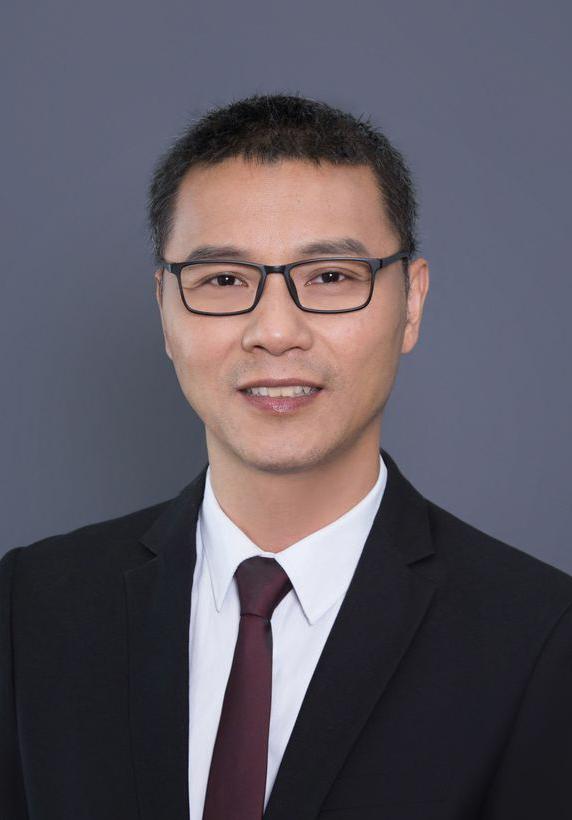}}]{Prof. S. Kevin Zhou} (Fellow, IEEE) obtained his Ph.D degree from the University of Maryland, College Park. Currently, he is a distinguished professor and founding executive dean of the School of Biomedical Engineering, Suzhou Institute for Advanced Research, University of Science and Technology of China (USTC), and an adjunct professor at the Institute of Computing Technology, Chinese Academy of Sciences. He directs the Center for Medical Imaging, Robotics, Analytic Computing and Learning (MIRACLE). Prior to this, he was a principal expert and a senior R\&D director at Siemens Healthcare Research. Dr. Zhou has published 260+ book chapters and peer-reviewed journal and conference papers, registered 140+ granted patents, written three research monographs, and edited three books. The most recent book he led the edition is entitled ``Handbook of Medical Image Computing and Computer Assisted Intervention, SK Zhou, D Rueckert, G Fichtinger (Eds.)" and the book he coauthored most recently is entitled ``Deep Network Design for Medical Image Computing, H Liao, SK Zhou, J Luo". He has won multiple awards including R\&D 100 Award (Oscar of Invention), Siemens Inventor of the Year, UMD ECE Distinguished Alumni Award, BMEF Editor of the Year, and Finalist Paper for MICCAI Young Scientist Award (twice). He has been a program co-chair for MICCAI2020, and an associate editor for IEEE Trans. Medical Imaging, IEEE Trans. Pattern Analysis Machine Intelligence, Medical Image Analysis, and an area chair for AAAI, CVPR, ICCV, MICCAI, and NeurIPS. He has been elected as a treasurer and board member of the MICCAI Society, an advisory board member of MONAI (Medical Open Network for AI), and a fellow of AIMBE, IAMBE, IEEE, MICCAI, and NAI.
\end{IEEEbiography}

\end{document}